\documentclass[preprint,12pt,authoryear]{elsarticle}

\usepackage{amssymb}
\usepackage{amsmath}
\usepackage{amsthm}

\newtheorem{theorem}{Theorem}
\newtheorem{corollary}{Corollary}[theorem] 

\usepackage{algorithm}
\usepackage{booktabs}
\usepackage{algpseudocode}
\usepackage{makecell}
\usepackage{hyperref}
\usepackage{etoolbox}

\journal{Medical Image Analysis}

\begin{document}

\begin{frontmatter}



\title{Projection Embedded Diffusion Bridge for CT Reconstruction from Incomplete Data} 


\author[label1,label2]{Yuang Wang} 
\author[label2]{Pengfei Jin}
\author[label2]{Siyeop Yoon}
\author[label2]{Matthew Tivnan}
\author[label1]{Shaoyang Zhang}
\author[label1]{Li Zhang}
\author[label2]{Quanzheng Li}
\author{Zhiqiang Chen\corref{cor1}\fnref{label1}}
\ead{czq@mail.tsinghua.edu.cn}
\author[label2,label3]{Dufan Wu\corref{cor1}}
\cortext[cor1]{Corresponding authors.}
\ead{dufan.wu@osumc.edu}
\affiliation[label1]{organization={The Department of Engineering Physics, Tsinghua University},
            addressline={30 Shuangqing Road, Haidian}, 
            city={Beijing},
            postcode={100084},
            country={China}}
\affiliation[label2]{organization={Center for Advanced Medical Computing and Analysis, Massachusetts General Hospital and Harvard Medical School},
            addressline={399 Revolution Dr}, 
            city={Somerville},
            postcode={02145}, 
            state={Massachusetts},
            country={USA}}
\affiliation[label3]{organization={Department of Radiology, The Ohio State University},
            addressline={2050 Kenny Rd}, 
            city={Columbus},
            postcode={OH 43221}, 
            state={Ohio},
            country={USA}}

\begin{abstract}
Reconstructing CT images from incomplete projection data remains challenging due to the ill-posed nature of the problem. Diffusion bridge models have recently shown promise in restoring clean images from their corresponding Filtered Back Projection (FBP) reconstructions, but incorporating data consistency into these models remains largely underexplored. Incorporating data consistency can improve reconstruction fidelity by aligning the reconstructed image with the observed projection data, and can enhance detail recovery by integrating structural information contained in the projections. In this work, we propose the Projection Embedded Diffusion Bridge (PEDB). PEDB introduces a novel reverse stochastic differential equation (SDE) to sample from the distribution of clean images conditioned on both the FBP reconstruction and the incomplete projection data. By explicitly conditioning on the projection data in sampling the clean images, PEDB naturally incorporates data consistency. We embed the projection data into the score function of the reverse SDE. Under certain assumptions, we derive a tractable expression for the posterior score. In addition, we introduce a free parameter to control the level of stochasticity in the reverse process. We also design a discretization scheme for the reverse SDE to mitigate discretization error. Extensive experiments demonstrate that PEDB achieves strong performance in CT reconstruction from three types of incomplete data, including sparse-view, limited-angle, and truncated projections. For each of these types, PEDB outperforms evaluated state-of-the-art diffusion bridge models across standard, noisy, and domain-shift evaluations.
\end{abstract}



\begin{keyword}
CT reconstruction \sep incomplete data \sep diffusion bridge \sep data consistency


\end{keyword}

\end{frontmatter}



\section{Introduction}
Computed Tomography (CT) plays a vital role in medical imaging by delivering high-resolution cross-sectional images of internal anatomical structures. However, practical constraints during data acquisition often result in incomplete projections, giving rise to multiple ill-posed inverse problems. In sparse-view CT, the number of projection angles is significantly reduced, typically for faster sampling, leading to larger angular intervals and substantial loss of information. In limited-angle CT, projections are acquired over a restricted angular range, often due to mechanical limitations or clinical requirements, causing pronounced directional artifacts. In cases involving truncated projections, parts of the projection data are missing because of limited detector size or a constrained field of view (FOV), as commonly encountered in large-body imaging or region-of-interest scans.  These challenging scenarios motivate the development of advanced reconstruction approaches that can reliably recover high-fidelity images from incomplete data.

Score-based diffusion models~\citep{ho2020denoising,chung2022diffusion} have achieved strong performance in CT reconstruction from incomplete data. However, their reverse process is typically initialized from pure Gaussian noise, which contains no structural information about the clean image distribution and may therefore be suboptimal. Diffusion bridge models~\citep{liu20232,wang2025implicit} have recently emerged as a promising alternative. By learning a stochastic transformation between clean images and their corresponding Filtered Back Projection (FBP) reconstructions, they enable the reverse process to start directly from the FBP image. This initialization makes diffusion bridge models inherently better suited for CT reconstruction.

Incorporating data consistency into diffusion bridge models is both critical and challenging. It can improve reconstruction fidelity by aligning the reconstructed image with the observed projection data, and can enhance detail
recovery by integrating structural information contained in the projections. While data consistency has been extensively studied in diffusion models, it remains largely underexplored in diffusion bridges. Representative diffusion bridge models, such as Image-to-Image Schrödinger Bridge (I$^2$SB)~\citep{liu20232} and Denoising Diffusion Bridge Model (DDBM)~\citep{zhou2023denoising}, operate entirely in the image domain and do not explicitly incorporate data consistency.  Consistent Direct Diffusion Bridge (CDDB)~\citep{chung2024direct} introduces a single-step data consistency mechanism into I²SB, demonstrating effectiveness in natural image restoration. However, its direct application to CT reconstruction from incomplete data offers limited performance gains, primarily due to the severely ill-posed nature of the CT system matrix.
 
In this work, we propose the Projection Embedded Diffusion Bridge (PEDB) for CT reconstruction from incomplete data. From image-domain diffusion bridges, we adopt the forward process to transition clean images into their corresponding FBP reconstructions and the trained data predictor to capture image-domain prior information. For the reverse process, we propose a novel reverse stochastic differential equation (SDE) to sample from the distribution of clean images conditioned on both the FBP reconstruction and the observed projection data. By explicitly conditioning on the projection data in sampling the clean images, PEDB naturally incorporates data consistency. We embed the projection data into the score function of the reverse SDE. Under certain assumptions, we derive a tractable expression for the posterior score. Additionally, we introduce a free parameter into the reverse SDE to control the level of stochasticity in the reverse process. We also design a discretization scheme for the reverse SDE to mitigate discretization error during sampling.

We evaluate the effectiveness of PEDB for CT reconstruction from three types of incomplete projection data,  including sparse-view, limited-angle, and truncated projections. For each type, we assess PEDB’s baseline performance through standard evaluations without added noise or domain shift, examine its performance at noise levels higher than that in training, and evaluate its generalization ability to unseen anatomical structures and to geometry-induced variations in FBP artifact patterns. Extensive experiments demonstrate PEDB’s strong performance in both quantitative metrics and visual quality. Across all three types of incomplete data, PEDB outperforms evaluated state-of-the-art diffusion bridge models in all evaluation scenarios. 

The main contributions of our work are fourfold:
\begin{itemize}
\item{We propose the Projection Embedded Diffusion Bridge (PEDB). To the best of our knowledge, PEDB is among the first approaches to incorporate data consistency into diffusion bridge models for CT reconstruction from incomplete data.}
\item{We establish the theoretical validity of the proposed reverse SDE for sampling from the distribution of clean images conditioned on both the FBP reconstruction and the incomplete projection data.}
\item{We derive a tractable expression for the posterior score under certain assumptions, and design a discretization scheme for the reverse SDE.}
\item{Through extensive experiments, we demonstrate PEDB's strong performance in CT reconstruction from three types of incomplete data, including sparse-view, limited-angle, and truncated projections. For each of these types, PEDB outperforms evaluated state-of-the-art diffusion bridge models across standard, noisy, and domain-shift evaluations.}

\end{itemize}
\section{Related Work}
In this section, we summarize representative methods for CT reconstruction from incomplete data, with particular emphasis on diffusion-based approaches. We also introduce recent developments in diffusion bridge models, describing their potential for CT reconstruction and the remaining challenges in incorporating data consistency.

\subsection{Methods for CT Reconstruction from Incomplete Data}
In CT reconstruction from incomplete data, analytical methods such as FBP are often inadequate, producing severe artifacts. Preprocessing strategies reduce artifacts by modifying the projection data before FBP reconstruction, such as Riess reweighting~\citep{riess2013tv} for limited-angle projections and smooth projection extrapolation~\citep{hsieh2004novel} for truncated data. Iterative reconstruction methods instead formulate the task as an optimization problem that combines a data consistency term with handcrafted regularization. Commonly used regularizers include total variation (TV)~\citep{yu2009compressed,vandeghinste2011split} and directional TV (DTV)~\citep{zhang2021directional,chen2024prototyping}, which reduces artifacts by encouraging sparsity in image gradients. 

Deep learning has significantly advanced CT reconstruction from incomplete data. One line of research focuses on developing end-to-end mapping networks that operate in the projection-domain, the image-domain, or both. Projection-domain networks~\citep{8513659,8452958} learn to complete missing projection data before FBP reconstruction; image-domain networks~\citep{7949028,8331861} learn a direct mapping from FBP reconstructions to clean images; and dual-domain networks~\citep{ketola2021generative,ZHOU2022102289,LI2025103420} combine both strategies by jointly training projection- and image-domain subnetworks under a dual-domain loss, enabling information exchange across domains. These networks are typically trained using fidelity losses (e.g., Mean Square Error (MSE)) and perceptual losses (e.g., Learned Perceptual Image Patch Similarity (LPIPS)~\citep{zhang2018perceptual}). Additionally, adversarial losses from generative adversarial networks (GANs)~\citep{10.5555/2969033.2969125} have been incorporated to enhance detail restoration~\citep{8513659,ketola2021generative}. 
Another line of research integrates deep learning with iterative reconstruction. Plug-and-play methods~\citep{gupta2018cnn,wang2019admm} employ pretrained image-domain networks as learned priors to regularize the iterative reconstruction process. Unrolled networks~\citep{8290981,https://doi.org/10.1002/mp.13627} unfold iterative reconstruction into a deep network, explicitly incorporating data consistency and replacing handcrafted regularization with trainable network components. In addition to these two lines, score-based diffusion models have demonstrated strong performance in CT reconstruction from incomplete data. These works are summarized in subsection~\ref{sec: relat_diffusion}.

\subsection{Diffusion Models for CT Reconstruction from Incomplete Data}
\label{sec: relat_diffusion}
Diffusion models have shown strong performance in CT reconstruction from incomplete data and are commonly grouped into unsupervised and supervised approaches. Unsupervised models learn unconditional score functions as priors for the data distribution and incorporate data consistency during inference, allowing a single trained network to be applied across different types of incomplete data. Solving Inverse Problems with Score-based Generative models (SIM-SGM)~\citep{song2022solving} incorporates data consistency into intermediate images via proximal optimization. Manifold Constrained Gradient (MCG)~\citep{chung2022improving} introduces a gradient correction term to ensure that intermediate images remain on the true noisy manifold after data consistency is applied. Diffusion Posterior Sampling (DPS)~\citep{chung2022diffusion} derives a tractable approximation of the projection-conditioned score function under certain assumptions, enabling sampling from the distribution of clean images conditioned on incomplete data. Subsequent studies~\citep{10377891,10.1007/978-3-031-72744-3_7} have demonstrated the effectiveness of DPS in sparse-view CT reconstruction. Decomposed Diffusion Sampling (DDS)~\citep{chung2024decomposed} imposes conditions on the data manifold and eliminates the need for computing the Jacobian, which is required in MCG and DPS and is computationally expensive. ReSample~\citep{song2024solving} enforces hard data consistency on the latent-space expected mean and introduces a stochastic resampling strategy for updating intermediate latent states. More recently, Variational Score Solver (VSS)~\citep{he2024solving} integrates a latent diffusion model as a generative prior within the iterative reconstruction framework for regularization.

Supervised models instead learn conditional score functions, typically using the FBP reconstruction as the conditioning input. These models require paired clean images and FBP reconstructions for training and must be retrained for each type of incomplete data, but they often achieve higher reconstruction quality than unsupervised approaches. Diffusion Probabilistic Limited-Angle CT Reconstruction (DOLCE)~\citep{liu2023dolce} employs a conditional denoising diffusion probabilistic model (cDDPM)~\citep{ho2020denoising} and incorporates data consistency into intermediate images during the reverse process, similar in spirit to SIM-SGM. Cascaded Multi-path Shortcut Diffusion Model (CMDM)~\citep{zhou2024cascaded} also adopts a cDDPM and employs a jump-start strategy~\citep{meng2021sdedit} to initialize the reverse process from a GAN-based reconstruction with added Gaussian noise. More recently, Prior-Guided and Wavelet Enhanced Diffusion Model (PWD)~\citep{liu2025pwd} employs a conditional denoising diffusion implicit model (cDDIM)~\citep{song2020denoising} and incorporates additional image-domain guidance from the FBP reconstruction.

\subsection{Diffusion Bridge Models}
\label{sec: relat_bridge}
Diffusion bridge models have recently emerged as a promising alternative to diffusion models for image reconstruction and restoration tasks. These models establish a stochastic transformation between paired clean and corrupted images. In the forward process, intermediate images can be sampled analytically, allowing the data predictor to be trained with efficiency comparable to diffusion models. In the reverse process, unlike diffusion models that start from pure Gaussian noise, diffusion bridges begin from the corrupted image, which already contains substantial structural information.

Most existing diffusion bridges do not explicitly incorporate data consistency; we refer to these as image-domain diffusion bridges. To sample from the distribution of clean images conditioned on a given corrupted image, these models propose different reverse sampling strategies. For example, Inversion by Direct Iteration (InDI)~\citep{delbracio2023inversion} and  I$^2$SB~\citep{liu20232} develop Markovian samplers; Implicit Image-to-Image Schrödinger Bridge (I$^3$SB)~\citep{wang2025implicit}, Diffusion Bridge Implicit Model (DBIM)~\citep{zheng2025diffusion} and Brownian Bridge~\citep{wang2024score} introduce non-Markovian samplers; and DDBM~\citep{zhou2023denoising} presents a hybrid sampler that alternates between a reverse SDE and a probability flow ordinary differential equation (ODE). These image-domain diffusion bridges have outperformed diffusion models in natural image restoration tasks~\citep{liu20232}. In CT reconstruction from incomplete data, their application remains limited, but they have shown strong potential. In particular, I$^2$SB and I$^3$SB demonstrate superior perceptual quality in sparse view CT reconstruction compared to cDDPM and cDDIM~\citep{wang2025implicit}. 

While incorporating data consistency has been extensively studied in diffusion models, it remains largely unexplored in diffusion bridges. The only representative approach in natural image restoration is CDDB~\citep{chung2024direct} that integrates a single-step data consistency mechanism into I$^2$SB using principles from DDS. To the best of our knowledge, no prior work has incorporated data consistency into diffusion bridge models for CT reconstruction from incomplete data. Moreover, due to the ill-posed nature of the CT system matrix, directly applying CDDB to CT reconstruction yields only limited performance improvement. In this work, we propose PEDB for CT reconstruction from incomplete data. PEDB introduces projection data as an explicit condition when sampling the clean images in the reverse process, naturally incorporating data consistency.

\section{Preliminaries}
 This section presents the necessary preliminaries, including image-domain diffusion bridges, the data model that relates incomplete projections to clean images, and a standard result from Bayes’ theorem for Gaussian variables.

\subsection{Image-Domain Diffusion Bridge}
\label{sec: image domain diffusion bridge}
Image-domain diffusion bridges are designed to sample from the distribution of clean images conditioned on a given corrupted image. As shown in Consistency Diffusion Bridge Model (CDBM)~\citep{he2024consistency}, representative image-domain diffusion bridges, including the Brownian Bridge~\citep{wang2024score}, I$^2$SB~\citep{liu20232}, and DDBM~\citep{zhou2023denoising}, can be unified within the DDBM framework. Each model corresponds to a particular specification of the framework’s design space. In this subsection, we present the DDBM framework, including its forward process, data predictor, and reverse process. Since our focus is on the variance exploding (VE) schedule, the resulting formulations are simplified, with explanations for the simplifications provided in \ref{sec:explain preliminary}.
\subsubsection{Forward Process}
For CT reconstruction from incomplete data, the corrupted image is taken as the FBP reconstruction from incomplete projections, denoted by $X_{\text{FBP}}$. Given $X_{\text{FBP}}$, the forward process in VE diffusion bridges starts from the corresponding clean image $X_0$ at time $t=0$ and evolves forward in time according to the following forward SDE:
\begin{equation}
\text{d}X_t=g^2\left(t\right)h_t\left(X_t,X_{\text{FBP}}\right)\text{d}t+g\left(t\right)\text{d}w,
\label{eq:forward_sde}
\end{equation}
where $X_t\in \mathbb{R}^d$ is a continuous-time stochastic process over $t \in [0,T]$, $g\left(t\right)$ is the diffusion coefficient of $X_t$, $w$ denotes the standard Wiener process, and 
\begin{equation}
h_t\left(X_t,X_{\text{FBP}}\right)=\frac{1}{\overline{\sigma}_t^2}\left(X_{\text{FBP}}-X_t\right),
\label{eq: h}
\end{equation}
with $\overline{\sigma}_t^2=\int_t^Tg^2\left(\tau\right)\text{d}\tau$. Under this formulation, the forward process almost surely converges to  $X_\text{FBP}$ at time $t=T$:
\begin{equation}
q\left(X_T|X_{\text{FBP}}\right)=\delta\left(X_T-X_\text{FBP}\right),
\label{eq:q_T}
\end{equation}
where $\delta$ represents the Dirac function. Additionally, this forward process enables efficient analytical forward sampling. Given $X_0$ and $X_{\text{FBP}}$, the intermediate image $X_t$ follows a Gaussian distribution:
\begin{equation}
q\left(X_t|X_0,X_\text{FBP}\right)=\mathcal N\left(X_t;\frac{\overline{\sigma}^2_t}{\sigma^2_T}X_0+\frac{\sigma^2_t}{\sigma^2_T}X_{\text{FBP}},\frac{\sigma^2_t\overline{\sigma}^2_t}{\sigma^2_T}I\right),
\label{eq:forward sample}
\end{equation}
where $\sigma_t^2=\int_0^tg^2\left(\tau\right)\text{d}\tau$ and $\sigma_T^2=\sigma_t^2+\overline{\sigma}_t^2$.
\subsubsection{Image-Domain Data Predictor}
The image-domain data predictor $D_\theta\left(X_t,t,X_{\text{FBP}}\right)$ is trained by minimizing the following loss:
\begin{equation}
\theta^*=\arg\min_{\theta}\mathbb{E}_{X_0, X_{\text{FBP}}, t}\mathbb{E}_
{X_t\sim q\left(X_t|X_0,X_\text{FBP}\right)}\left[\lambda\left(t\right)\Vert D_{\theta}\left(X_t,t,X_\text{FBP}\right)-X_0 \Vert_2^2\right],
\label{eq:loss}
\end{equation}
where $X_0$ and $X_{\text{FBP}}$ are paired clean and corrupted training images, and $\lambda\left(t\right)>0$ is a weighting function. The image-domain expected mean $\hat{X}_{0}^{\left(t\right)}$ at time $t$ is defined as the conditional expectation of the clean image $X_0$ given $X_t$ and $X_{\text{FBP}}$: 
\begin{equation}
\label{eq: x0_hat}
\hat{X}_0^{\left(t\right)}=\mathbb{E}_{X_0\sim q_{\text{data}}\left(X_0|X_t,X_{\text{FBP}}\right)}\left[X_0\right].
\end{equation}
This expected mean can be predicted by the trained image-domain data predictor $D_{\theta^*}\left(X_t,t,X_\text{FBP}\right)$ as:
\begin{equation}
\hat{X}_{0}^{\left(t\right)}=D_{\theta^*}\left(X_t,t,X_\text{FBP}\right).
\label{eq:image_domain_expected_mean}
\end{equation}
\subsubsection{Reverse Process}
In the reverse process, image-domain diffusion bridges aim to sample from $q_{\text{data}}\left(X_0|X_{\text{FBP}}\right)$, the distribution of clean images conditioned solely on the corrupted image $X_{\text{FBP}}$. Sampling from this distribution can be achieved by initializing the reverse process with $X_T=X_{\text{FBP}}$ and simulating backward in time according to the reverse SDE:
\begin{equation}
\text{d}X_t=g^2\left(t\right)\left(h_t\left(X_t,X_{\text{FBP}}\right)-\nabla_{X_t}\log q\left(X_t|X_{\text{FBP}}\right)\right)\text{d}t+g\left(t\right)\text{d}\overline{w},
\label{eq: image domain reverse sde}
\end{equation}
where $\overline{w}$ denotes the reverse-time standard Wiener process. The score function $\nabla_{X_t}\log q\left(X_t|X_{\text{FBP}}\right)$ admits the following tractable form:
\begin{equation}
\nabla_{X_t}\log q\left(X_t|X_{\text{FBP}}\right)=\frac{1}{\overline{\sigma}_t^2}X_{\text{FBP}}-\frac{\sigma_T^2}{\sigma_t^2\overline{\sigma}_t^2}X_t+\frac{1}{\sigma_t^2}\hat{X}_{0}^{\left(t\right)},
\label{eq: score matching}
\end{equation} 
where the image-domain expected mean  $\hat{X}_0^{\left(t\right)}$ is available from equation~(\ref{eq:image_domain_expected_mean}). 

\subsection{Data Model}
Instead of sampling from $q_{\text{data}}\left(X_0|X_{\text{FBP}}\right)$, which is the objective in image-domain diffusion bridges, we aim to sample from  $q_{\text{data}}\left(X_0|X_{\text{FBP}},y\right)$, where $y$ denotes the incomplete projection data and serves as an additional condition in sampling the clean images. The relationship between $y$ and the clean image $X_0$ is described by the following data model:
\begin{equation}
y=AX_0+n,
\end{equation}
where 
$A$ is the system matrix corresponding to the incomplete projection geometry, typically large and ill-posed, and 
$n$ represents measurement noise.
\subsection{Bayes' Theorem for Gaussian Variables}
\label{sec: Gaussian}
In our derivation, we make use of a standard result from Bayes' theorem for Gaussian variables~\citep{bishop2006pattern}. Specifically, suppose that a random variable $z_1$ follows a Gaussian distribution:
\begin{equation}
z_1\sim\mathcal N\left(z_1; \mu_1,\Lambda^{-1}\right),
\label{eq: z1}
\end{equation}
and that, conditioned on $z_1$, another variable $z_2$ also follows a Gaussian distribution:
\begin{equation}
z_2|z_1\sim\mathcal N\left(z_2; Mz_1+\mu_2,L^{-1}\right),
\label{eq: z2}
\end{equation}
then the conditional distribution of $z_1$ given $z_2$ is Gaussian as well, with mean 
\begin{equation}
\mathbb{E}\left[z_1|z_2\right]=\left(\Lambda+M^\mathsf{T}LM\right)^{-1}\left(M^{\mathsf{T}}L\left(z_2-\mu_2\right)+\Lambda\mu_1\right),
\label{eq: posterior_mean}
\end{equation}
and covariance
\begin{equation}
\text{cov}
\left[z_1|z_2\right]=\left(\Lambda+M^\mathsf{T}LM\right)^{-1}.
\label{eq: posterior_var}
\end{equation}

\section{Method}
\begin{figure*}[!t]
\centering
\includegraphics[width=\textwidth]{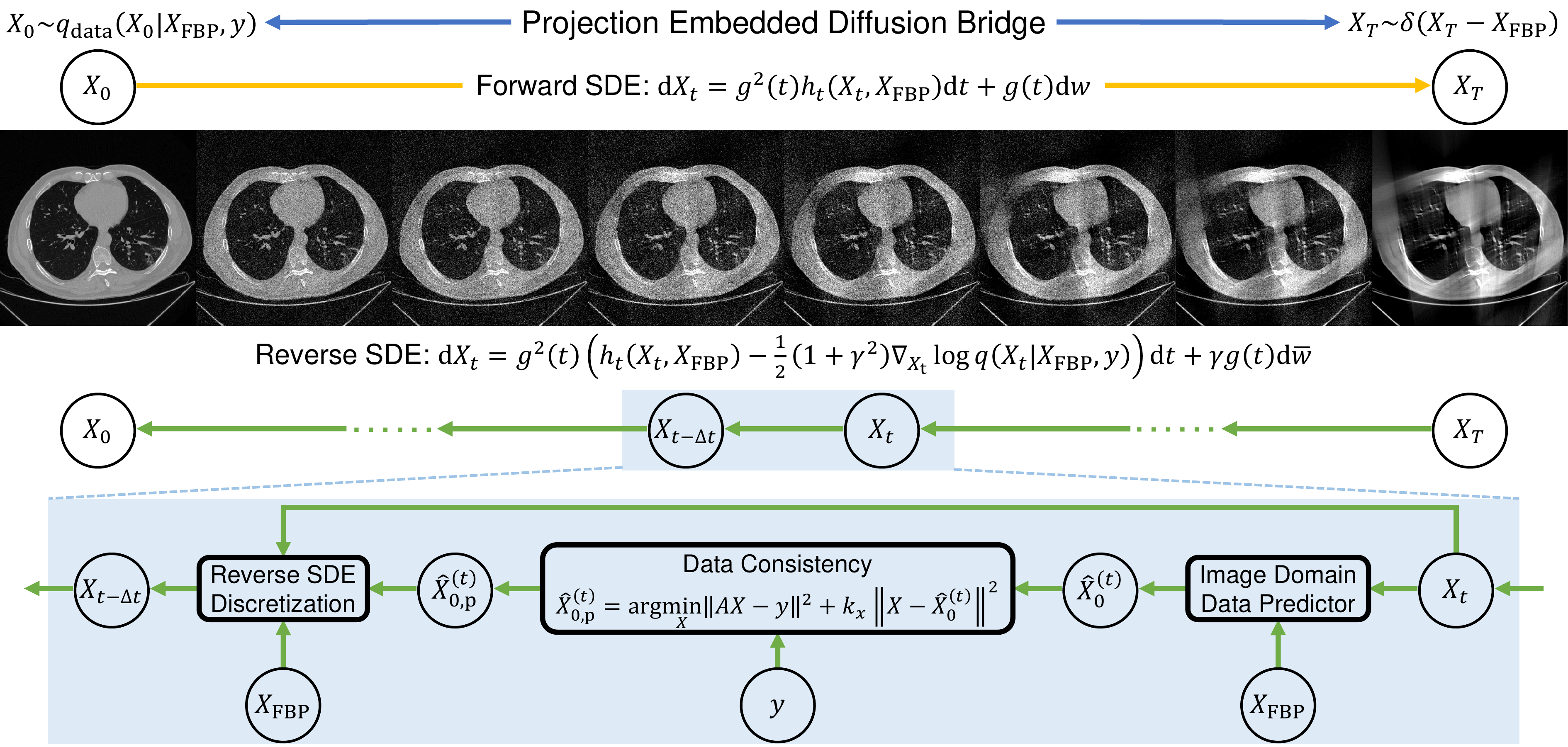}
\caption{Framework of the proposed PEDB. PEDB establishes a stochastic transformation between $q_{\text{data}}\left(X_0|X_{\text{FBP}},y\right)$ and $\delta\left(X_T-X_{\text{FBP}}\right)$. The forward process follows the same forward SDE as image-domain diffusion bridge models. In the reverse process, we propose a novel reverse SDE in which the incomplete projection data $y$ is embedded into the score function, and a free parameter $\gamma$ is introduced to control the level of stochasticity. At each reverse step from $t$ to $t-\Delta t$, the trained image-domain data predictor is used to predict the image-domain expected-mean $\hat{X}_0^{\left(t\right)}$. Data consistency is then incorporated by solving a quadratic optimization, producing the projection-embedded expected mean $\hat{X}_{0,\text{p}}^{\left(t\right)}$ from $\hat{X}_0^{\left(t\right)}$ and $y$. Using $\hat{X}_{0,\text{p}}^{\left(t\right)}$, $X_t$ and $X_{\text{FBP}}$, the next intermediate image $X_{t-\Delta t}$ is computed by discretizing the reverse SDE.}
\label{fig1}
\end{figure*}
In this section, we present our proposed method, Projection Embedded Diffusion Bridge (PEDB), for CT reconstruction from incomplete data. PEDB is designed to sample from $q_{\text{data}}\left(X_0|X_{\text{FBP}},y\right)$, the distribution of clean images conditioned on both the FBP reconstruction $X_{\text{FBP}}$ and the incomplete projection data $y$. Compared to image-domain diffusion bridges, PEDB incorporates $y$ as an additional condition in sampling the clean images. The motivation of this design is to incorporate data consistency, thereby enhancing reconstruction fidelity and improving detail recovery.

The overall framework of PEDB is illustrated in Figure~\ref{fig1}. PEDB establishes a stochastic transformation between $q_{\text{data}}\left(X_0|X_{\text{FBP}},y\right)$ and $\delta\left(X_T-X_{\text{FBP}}\right)$. In the forward process, we adopt the forward SDE~(\ref{eq:forward_sde}) from image-domain diffusion bridge models to transition the clean image $X_0$ to its corresponding FBP reconstruction $X_{\text{FBP}}$. In the reverse process, we start from $X_{\text{FBP}}$ and propose a novel reverse SDE to sample from $q_{\text{data}}\left(X_0|X_{\text{FBP}},y\right)$. In this reverse SDE, the projection data $y$ are embedded into the score function, formulating the posterior score $\nabla_{X_t}\log q\left(X_t|X_{\text{FBP}},y\right)$, and a free parameter $\gamma$ is introduced to control the level of stochasticity during sampling.

The main challenges of our approach are: (1) establishing the theoretical validity of the proposed reverse SDE for sampling from $q_{\text{data}}\left(X_0|X_{\text{FBP}},y\right)$; (2) deriving a tractable expression for the posterior score function with the image-domain expected mean $\hat{X}_0^{\left(t\right)}$ available from equation~(\ref{eq:image_domain_expected_mean}); and (3) designing a discretization scheme for the proposed reverse SDE to eliminate discretization error in both the linear drift and stochastic noise terms. In the following subsections, we first present the proposed reverse SDE, establish its theoretical validity, and explain the role of the free parameter $\gamma$; next, we derive a tractable expression for the posterior score under certain Gaussian assumptions; and finally, we describe our discretization strategy and present the complete sampling algorithm.
\subsection{Reverse SDE for PEDB}
\label{section: reverse sde}
In the reverse process, we propose initializing with $X_{T}=X_{\text{FBP}}$ and simulating backward in time using the following reverse SDE:
\begin{equation}
\text{d}X_t=g^2\left(t\right)\left(h_t\left(X_t,X_{\text{FBP}}\right)-\frac{1}{2}\left(1+\gamma^2\right)\nabla_{X_t}\log q\left(X_t|X_{\text{FBP}},y\right)\right)\text{d}t+\gamma g\left(t\right)\text{d}\overline{w},
\label{eq: reverse sde}
\end{equation}
where $\nabla_{X_t}\log q\left(X_t|X_{\text{FBP}},y\right)$ is the posterior score function with incomplete projection data $y$ embedded, and $\gamma\geq0$ is a free parameter that controls the level of stochasticity in the reverse process. 

Crucially, when both $X_{\text{FBP}}$ and $y$ are given, the reverse process described by the reverse SDE~(\ref{eq: reverse sde}) exactly inverts the forward process described by the forward SDE~(\ref{eq:forward_sde}), as both processes share the same marginal distributions at all time points.  This relationship is formalized in Theorem~\ref{theorem1}. A direct implication, presented in Corollary~\ref{corollary1}, is that simulating the proposed reverse SDE~(\ref{eq: reverse sde}) from $X_T=X_{\text{FBP}}$ backward in time from $T$ to 0 yields samples from $q_{\text{data}}\left(X_0|X_{\text{FBP}},y\right)$. Detailed proofs of Theorem~\ref{theorem1} and Corollary~\ref{corollary1} are provided in \ref{sec: proof_theorem1}.
\begin{theorem}
The reverse process described by the reverse SDE~(\ref{eq: reverse sde}) shares the same marginal distributions as the forward process described by the forward SDE~(\ref{eq:forward_sde}). Specifically, when both $X_{\text{FBP}}$ and $y$ are given, for any time $t\in[0,T]$, the intermediate image $X_t$ obtained by simulating the reverse SDE~(\ref{eq: reverse sde}) from $X_T=X_{\text{FBP}}$ backward to time $t$  follows the same marginal distribution $q\left(X_t|X_{\text{FBP}},y\right)$ as the intermediate image $X_t$ obtained by simulating the forward SDE~(\ref{eq:forward_sde}) from $X_0\sim q_{\text{data}}\left(X_0|X_{\text{FBP}},y\right)$ forward to time $t$.
\label{theorem1}
\end{theorem}
\begin{corollary}
Given both $X_{\text{FBP}}$ and $y$, initializing with $X_{T}=X_{\text{FBP}}$ and simulating the reverse SDE~(\ref{eq: reverse sde}) backward in time from $T$ to $0$ yields samples from $q_{\text{data}}\left(X_0|X_{\text{FBP}},y\right)$.
\label{corollary1}
\end{corollary}

To better understand the role of the free parameter $\gamma$, we decompose the proposed reverse SDE~(\ref{eq: reverse sde}) into two components, inspired by the formulation in Elucidating the Design Space of Diffusion-Based Generative Models (EDM)~\citep{Karras2022edm}. Specifically, the reverse SDE can be expressed as the sum of a probability flow ODE and a time-varying Langevin diffusion SDE:
\begin{equation}
\begin{aligned}
\text{d}X_t=&\underbrace{g^2\left(t\right)\left(h_t\left(X_t,X_{\text{FBP}}\right)-\frac{1}{2}\nabla_{X_t}\log q\left(X_t|X_{\text{FBP}},y\right)\right)\text{d}t}_{\text{probability flow ODE}}\\&\underbrace{-\frac{1}{2}\gamma^2g^2\left(t\right)\nabla_{X_t}\log q\left(X_t|X_{\text{FBP}},y\right)\text{d}t+\gamma g\left(t\right)\text{d}\overline{w}}_{\text{Langevin diffusion SDE}}.
\end{aligned}
\end{equation}
The probability flow ODE governs the deterministic evolution of $X_t$ from time $t$ to $t-\text{d}t$, ensuring a smooth progression along the diffusion trajectory.
The Langevin diffusion SDE, comprising both score-based correction and stochastic noise, encourages $X_t$ to align with the target marginal distribution $q\left(X_t|X_{\text{FBP}},y\right)$, with its intensity controlled by the free parameter $\gamma$. Under domain shift, where $X_t$ may deviate from the target marginal, the Langevin term helps pull it back toward the desired distribution. Thus, a larger $\gamma$, corresponding to a stronger contribution from the Langevin term, is expected to enhance PEDB’s generalization ability under domain shift. This effect is empirically supported by the ablation study presented in Subsection~\ref{sec: ablation_gamma}.
\subsection{Posterior Score Function}
To simulate the proposed reverse SDE~(\ref{eq: reverse sde}) backward in time, we require a tractable expression for the posterior score function $\nabla_{X_t}\log q\left(X_t|X_{\text{FBP}},y\right)$. In this subsection, we first establish its connection to a projection-embedded expected mean. Then, under certain Gaussian assumptions, we derive an expression that relates this expected mean to the image-domain expected mean $\hat{X}_0^{(t)}$ and the projection data $y$, thereby making the posterior score function tractable.
\subsubsection{Connection to Projection-Embedded Expected Mean}
\label{section: data consistency mean}

We define the projection-embedded expected mean $\hat{X}_{0,\text{p}}^{(t)}$ at time $t$ as the conditional expectation of the clean image $X_0$ given $X_t$, $X_{\text{FBP}}$ and the incomplete projection data $y$:
\begin{equation}
\label{eq: x0c}
\hat{X}_{0,\text{p}}^{\left(t\right)}=\mathbb{E}_{X_0\sim q_{\text{data}}\left(X_0|X_t,X_{\text{FBP}},y\right)}\left[X_0\right].
\end{equation}
Here, $y$ is incorporated as an additional conditioning variable, alongside $X_t$ and $X_{\text{FBP}}$, in modeling the clean image distribution. The posterior score function $\nabla_{X_t} \log q(X_t | X_{\text{FBP}}, y)$ is related to $\hat{X}_{0,\text{p}}^{(t)}$ via:
\begin{equation}
\nabla_{X_t}\log q\left(X_t|X_{\text{FBP}},y\right)=\frac{1}{\overline{\sigma}_t^2}X_{\text{FBP}}-\frac{\sigma_T^2}{\sigma_t^2\overline{\sigma}_t^2}X_t+\frac{1}{\sigma_t^2}\hat{X}_{0,\text{p}}^{\left(t\right)}.
\label{eq:score}
\end{equation}
A detailed derivation of equation~(\ref{eq:score}) is provided in \ref{sec: direvation_score}.

The posterior score $\nabla_{X_t}\log q\left(X_t|X_{\text{FBP}},y\right)$ shares a similar form with the original score $\nabla_{X_t}\log q\left(X_t|X_{\text{FBP}}\right)$ presented in equation~(\ref{eq: score matching}). The key difference lies in the replacement of the image-domain expected mean $\hat{X}_{0}^{\left(t\right)}$ with the projection-embedded expected mean $\hat{X}_{0,\text{p}}^{\left(t\right)}$.
Unlike $\hat{X}_{0}^{\left(t\right)}$ that is predicted by the trained predictor $D_{\theta^*}\left(X_t,t,X_\text{FBP}\right)$ as in equation·(\ref{eq:image_domain_expected_mean}), $\hat{X}_{0,\text{p}}^{\left(t\right)}$ is generally intractable without further assumptions.
\subsubsection{Gaussian Assumptions}
\label{sec: Gaussian_assumption}
 To obtain a computable form of $\hat{X}_{0,\text{p}}^{\left(t\right)}$, we introduce Gaussian assumptions on the FBP-conditioned data distribution $q_{\text{data}}\left(X_0|X_{\text{FBP}}\right)$ and the projection data likelihood $q\left(y|X_0,X_{\text{FBP}}\right)$. Specifically, we assume that the clean image $X_0$, conditioned on $X_{\text{FBP}}$, follows a Gaussian distribution with mean $Z\left(X_{\text{FBP}}\right)$ and covariance $\sigma_x^2I$:
\begin{equation}
q_{\text{data}}\left(X_0|X_{\text{FBP}}\right)=\mathcal{N}\left(X_0;Z\left(X_{\text{FBP}}\right),\sigma_x^2I\right),
\label{eq: assume1}
\end{equation}
where $Z$ is an unknown function that estimates $X_0$ from $X_{\text{FBP}}$, and $\sigma_x^2$ reflects the uncertainty in this estimation. In addition, we assume that the projection data $y$, conditioned on both $X_0$ and $X_{\text{FBP}}$, follows a Gaussian distribution with mean $AX_0$ and covariance $\sigma_y^2I$:
\begin{equation}
q\left(y|X_0,X_{\text{FBP}}\right)=\mathcal{N}\left(y;AX_0,\sigma_y^2I\right).
\label{eq: assume_y}
\end{equation}
Under these Gaussian assumptions, the projection-embedded expected mean can be expressed in terms of the image-domain expected mean and the projection data, as detailed in Subsubsection \ref{section: tractable form}, which makes the posterior score function tractable.

\subsubsection{Tractable Form of Projection-Embedded Expected Mean}
\label{section: tractable form}
Our goal in this subsubsection  is to express the projection-embedded expected mean $\hat{X}_{0,\text{p}}^{(t)}$ in terms of the image-domain expected mean $\hat{X}_0^{(t)}$ and the observed projection data $y$. Since $\hat{X}_0^{(t)}$ and $\hat{X}_{0,\text{p}}^{(t)}$ are defined as the expectations of $q_{\text{data}}\left(X_0|X_t,X_\text{FBP}\right)$ and $q_{\text{data}}\left(X_0|X_t,X_\text{FBP},y\right)$, respectively, we can relate them by demonstrating that both distributions are Gaussian under the assumptions introduced in Subsubsection~\ref{sec: Gaussian_assumption}, and then applying equation~(\ref{eq: posterior_mean}) from the Bayes’ theorem for Gaussian variables presented in Subsection~\ref{sec: Gaussian}.

We begin by showing  that $q_{\text{data}}\left(X_0|X_t,X_\text{FBP}\right)$ is Gaussian. To apply Bayes' theorem for Gaussian variables, we identify $X_0$ as $z_1$ in equation~(\ref{eq: z1}), $X_t$ as $z_2$ in equation~(\ref{eq: z2}), and treat $X_\text{FBP}$ as a given condition. Since $q_\text{data}\left(X_0|X_{\text{FBP}}\right)$ is assumed to be Gaussian (equation~(\ref{eq: assume1})) and $q\left(X_t|X_0,X_{\text{FBP}}\right)$ is also Gaussian (equation~(\ref{eq:forward sample})), it follows that $q_{\text{data}}\left(X_0|X_t,X_\text{FBP}\right)$ is Gaussian as well:
\begin{equation}
q_{\text{data}}\left(X_0|X_t,X_\text{FBP}\right)=\mathcal{N}\left(X_0;\hat{X}_0^{\left(t\right)},\frac{\sigma_x^2\sigma_t^2\sigma_T^2}{\sigma_t^2\sigma_T^2+\overline{\sigma}_t^2\sigma_x^2}I\right),
\label{eq: 0|t_FBP}
\end{equation}
where the covariance is computed using equation~(\ref{eq: posterior_var}).

Next, we show that  $q_{\text{data}}\left(X_0|X_t,X_{\text{FBP}},y\right)$ is also Gaussian. To apply Bayes' theorem for Gaussian variables, we identify $X_0$ as $z_1$ in equation~(\ref{eq: z1}) and $y$ as $z_2$ in equation~(\ref{eq: z2}), with $X_t$ and $X_\text{FBP}$ treated as given conditions. The distribution $q_{\text{data}}\left(X_0|X_t,X_{\text{FBP}}\right)$ is already shown to be Gaussian in equation~(\ref{eq: 0|t_FBP}). To determine the form of  $q\left(y|X_0,X_t,X_{\text{FBP}}\right)$, we use the conditional independence between $y$ and $X_t$ given $X_0$ and $X_\text{FBP}$, which gives:
\begin{equation}
q\left(y|X_0,X_t,X_{\text{FBP}}\right)=q\left(y|X_0,X_{\text{FBP}}\right).
\end{equation}
The right-hand side is assumed to be Gaussian in equation~(\ref{eq: assume_y}), and therefore $q\left(y|X_0,X_t,X_{\text{FBP}}\right)$ is Gaussian as well. Since both $q_{\text{data}}\left(X_0|X_t,X_{\text{FBP}}\right)$ and $q\left(y|X_0,X_t,X_{\text{FBP}}\right)$ are Gaussian, it follows from Bayes' theorem for Gaussian variables that $q_{\text{data}}\left(X_0|X_t,X_{\text{FBP}},y\right)$ is also Gaussian. Using equation~(\ref{eq: posterior_mean}), its expectation $\hat{X}_{0,\text{p}}^{\left(t\right)}$ can be written as:
\begin{equation}
\hat{X}_{0,\text{p}}^{\left(t\right)}=\left(A^{\mathsf{T}}A+k_xI\right)^{-1}\left(A^\mathsf{T}y+k_x\hat{X}_0^{\left(t\right)}\right),
\label{eq: x0c_raw}
\end{equation}
where $k_x=\left(1+\frac{\overline{\sigma}_t^2\sigma_x^2}{\sigma_t^2\sigma_T^2}\right)\frac{\sigma_y^2}{\sigma_x^2}$. 
 
With equation~(\ref{eq: x0c_raw}), $\hat{X}_{0,\text{p}}^{\left(t\right)}$ becomes tractable, as it is explicitly expressed in terms of $\hat{X}_0^{\left(t\right)}$ and $y$. To better understand this expression, we reformulate $\hat{X}_{0,\text{p}}^{\left(t\right)}$ as the solution to the following optimization problem:
\begin{equation}
\hat{X}_{0,\text{p}}^{\left(t\right)}=\arg\min_{X}\Vert AX-y\Vert_2^2+k_x\Vert X-\hat{X}_{0}^{(t)}\Vert _2^2.
\label{eq:optimization}
\end{equation}
Since the gradient of the quadratic objective in equation~(\ref{eq:optimization}) vanishes at the point given by equation~(\ref{eq: x0c_raw}), the two expressions for $\hat{X}_{0,\text{p}}^{\left(t\right)}$ are equivalent. In this optimization problem, the hyperparameter $k_x$ balances the data consistency term $\Vert AX-y\Vert_2^2$ and the regularization term $\Vert X-\hat{X}_{0}^{(t)}\Vert _2^2$. In the range space of the system matrix $A$, the data consistency term enforces alignment of  $\hat{X}_{0,\text{p}}^{\left(t\right)}$  with the observed projection data $y$, and the regularization term suppresses the impact of measurement noise in the projection data by incorporating the image-domain expected mean $\hat{X}_{0}^{\left(t\right)}$. In the null space of $A$, where $y$ provides no information, the solution depends solely on the regularization term, and  $\hat{X}_{0,\text{p}}^{\left(t\right)}$ naturally reduces to $\hat{X}_{0}^{\left(t\right)}$, thereby preserving prior-driven information in the unconstrained subspace.

\subsubsection{Summary for Posterior Score Function}
Under the Gaussian assumptions in equations~(\ref{eq: assume1}) and~(\ref{eq: assume_y}), the posterior score function $\nabla_{X_t}\log q\left(X_t|X_{\text{FBP}},y\right)$ admits a tractable form. Specifically, with the image-domain expected mean $\hat{X}_0^{\left(t\right)}$ available from equation~(\ref{eq:image_domain_expected_mean}), the projection-embedded expected mean $\hat{X}_{0,\text{p}}^{(t)}$ can be obtained from $\hat{X}_0^{\left(t\right)}$ and the observed projection data $y$ via equation~(\ref{eq:optimization}). The posterior score $\nabla_{X_t}\log q\left(X_t|X_{\text{FBP}},y\right)$ can be then computed from $\hat{X}_{0,\text{p}}^{(t)}$  using equation~(\ref{eq:score}). This tractable formulation of the posterior score enables the simulation of the proposed reverse SDE~(\ref{eq: reverse sde}).
\subsection{Reverse SDE Discretization}
Directly discretizing the proposed reverse SDE~(\ref{eq: reverse sde}) using the Euler–Maruyama method~\citep{doi:10.1137/S0036144500378302} can lead to significant discretization error, especially when a large step size $\Delta t$ is used. To reduce this error, we reformulate the reverse SDE~(\ref{eq: reverse sde}) into the following equivalent form:
\begin{equation}
\label{eq: equivalent sde}
\text{d}\frac{X_t}{\sigma_t^{\left(1+\gamma^2\right)}\overline{\sigma}_t^{\left(1-\gamma^2\right)}}=\frac{X_{\text{FBP}}}{\sigma_T^2}\text{d}\left(\frac{\sigma_t}{\overline{\sigma}_t}\right)^{1-\gamma^2}+\frac{\hat{X}_{0,\text{p}}^{\left(t\right)}}{\sigma_T^2}\text{d}\left(\frac{\overline{\sigma}_t}{\sigma_t}\right)^{1+\gamma^2}+\frac{\gamma g\left(t\right)}{\sigma_t^{\left(1+\gamma^2\right)}\overline{\sigma}_t^{\left(1-\gamma^2\right)}}\text{d}\overline{w}.
\end{equation}
A detailed proof of the equivalence between the original reverse SDE~(\ref{eq: reverse sde}) and this reformulation in equation~(\ref{eq: equivalent sde}) is provided in \ref{sec: sde_equivalence}.

The key advantage of this reformulation is that both sides of equation~(\ref{eq: equivalent sde}) can be integrated analytically over the time interval $[t - \Delta t, t]$, assuming the projection-embedded expected mean $\hat{X}_{0,\text{p}}^{(t)}$ remains approximately constant within this short interval. This yields the following discretized update rule:
\begin{equation}
X_{t-\Delta t}=a_{t}\hat{X}_{0,\text{p}}^{\left(t\right)}+b_{t}X_{t}+c_{t}X_\text{FBP}+\eta_{t}\epsilon_{t},
\label{eq: discretize reverse sde}
\end{equation}
where $\epsilon_t \sim \mathcal{N}(0, I)$ is standard Gaussian noise. The coefficients $a_t$, $b_t$, and $c_t$ depend on the noise scaling factor $\eta_t$:
\begin{subequations}\label{eq:ABC}
\begin{align}
a_t&=\frac{\overline{\sigma}_{t-\Delta t}^2}{\sigma_T^2}-\frac{\overline{\sigma}_{t}^2}{\sigma_{T}^2}\frac{\sqrt{\sigma_{t-\Delta t}^2\overline{\sigma}_{t-\Delta t}^2-\eta_t^2\sigma_T^2}}{\sigma_{t}\overline{\sigma}_{t}},\\
b_t &= \frac{\sqrt{\sigma_{t-\Delta t}^2\overline{\sigma}_{t-\Delta t}^2-\eta_t^2\sigma_T^2}}{\sigma_t\overline{\sigma}_t},\\
c_t&=\frac{\sigma_{t-\Delta t}^2}{\sigma_T^2}-\frac{\sigma_t^2}{\sigma_{T}^2}\frac{\sqrt{\sigma_{t-\Delta t}^2\overline{\sigma}_{t-\Delta t}^2-\eta_t^2\sigma_T^2}}{\sigma_{t}\overline{\sigma}_{t}},
\end{align}
\end{subequations}
with $\eta_t$ itself determined by the free parameter $\gamma$:
\begin{equation}
\eta_t=\frac{\sigma_{t-\Delta t}\overline{\sigma}_{t-\Delta t}}{\sigma_T}\sqrt{1-\left(\frac{\sigma_{t-\Delta t}\overline{\sigma}_t}{\overline{\sigma}_{t-\Delta t}\sigma_t}\right)^{2\gamma^2}}.
\label{eq: eta_t}
\end{equation}
Detailed derivations of equations~(\ref{eq: discretize reverse sde}) through~(\ref{eq: eta_t}) are provided in ~\ref{sec: derive_discretize_sde}. In this scheme, discretization error arises solely from $\hat{X}_{0,\text{p}}^{(t)}$, whose nonlinearity with respect to $X_t$ prevents exact integration and motivates approximating it as constant within each step. The linear drift and stochastic noise terms are integrated analytically, ensuring they introduce no additional discretization error.

The proposed update rule resembles the non-Markovian update in I$^3$SB~\citep{wang2025implicit}, but with two key distinctions. First, it replaces the image-domain expected mean $\hat{X}_{0}^{(t)}$ with the projection-embedded expected mean $\hat{X}_{0,\text{p}}^{(t)}$, thereby explicitly incorporating data consistency into the reverse process. Second, rather than treating the noise scaling factor $\eta_t$ as a free hyperparameter, we formulate it in equation~(\ref{eq: eta_t}) as a function of the parameter $\gamma$, which controls the strength of the Langevin dynamics in the reverse SDE~(\ref{eq: reverse sde}). Because $0 \leq \frac{\sigma_{t-\Delta t}\overline{\sigma}_t}{\overline{\sigma}_{t-\Delta t}\sigma_t} < 1$, increasing $\gamma$ results in a larger $\eta_t$ and a smaller $b_t$, meaning that more of the deterministic noise in $X_t$ is replaced by newly injected stochastic noise. This provides a discretized perspective on how $\gamma$ controls the level of stochasticity in the reverse process.
\begin{algorithm}[t]
\footnotesize
\renewcommand{\algorithmicrequire}
   {\textbf{Input:}}
   \renewcommand{\algorithmicensure}
   {\textbf{Output:}}
\caption{Reverse Process of PEDB}\label{alg1}
\begin{algorithmic}
\Require incomplete data $y$, system matrix A, time schedule $\{t_i\}_{i=0}^{N}$,\\ \qquad $\text{ }$ trained image-domain data predictor $D_{\theta^*}\left(X_t,t,X_{\text{FBP}}\right)$ \\
\textbf{Initialize:} $X_{t_N}=X_T=X_{\text{FBP}}=\text{FBP}\left(y\right)$
\For{$n=N$ to $1$}   
\State
$\hat{X}_0^{\left(t_n\right)}= D_{\theta^*}\left(X_{t_n},t_n,X_{\text{FBP}}\right)$
\hfill\Comment{Image-domain expected mean}
\State
Start from $\hat{X}_0^{\left(t_n\right)}$, perform $m$-step CG to solve: 
\hfill\Comment{Data consistency}
\State\qquad
$\hat{X}_{0,\text{p}}^{\left(t_n\right)}=\arg\min_{X}\Vert AX-y\Vert_2^2+k_x\Vert X-\hat{X}_0^{\left(t_n\right)}\Vert _2^2$,
\State
$X_{t_{n-1}}=a_{t_n}\hat{X}_{0,\text{p}}^{\left(t_n\right)}+b_{t_n}X_{t_n}+c_{t_n}X_\text{FBP}+\eta_{t_n}\epsilon_{t_n}$
\hfill\Comment{Reverse SDE discretization}
\EndFor
\Ensure $X_0=X_{t_0}$
\end{algorithmic}
\end{algorithm}
\subsection{Algorithm}
The reverse process of PEDB is summarized in Algorithm~\ref{alg1}. The proposed method takes as input the incomplete projection data $y$, the system matrix $A$, the trained image-domain data predictor $D_{\theta^*}(X_t, t, X_{\text{FBP}})$, and a predefined time schedule $\{t_i\}_{i=0}^{N}$. The reverse process is initialized with the corrupted image $X_{\text{FBP}}$, reconstructed from $y$ using the FBP algorithm. At each reverse step, three operations are performed. First, the image-domain expected mean $\hat{X}_0^{(t)}$ is predicted using the trained predictor $D_{\theta^*}(X_t, t, X_{\text{FBP}})$ as described in equation~(\ref{eq:image_domain_expected_mean}). Second, data consistency is incorporated by solving the optimization problem in equation~(\ref{eq:optimization}), yielding the projection-embedded expected mean $\hat{X}_{0,\text{p}}^{(t)}$ from $\hat{X}_0^{(t)}$ and $y$. To improve computational efficiency, this quadratic optimization is initialized with $\hat{X}_0^{(t)}$ and approximated using an $m$-step Conjugate Gradient (CG) update~\citep{refsnaes2009brief}. Third, the next intermediate image is computed using the designed discretization scheme, combining $\hat{X}_{0,\text{p}}^{(t)}$, $X_t$, and $X_{\text{FBP}}$ along with added Gaussian noise, as described in equation~(\ref{eq: discretize reverse sde}).

\section{Experiments}
We conducted comprehensive experiments to evaluate PEDB in CT reconstruction from three representative types of incomplete projection data, including sparse-view, limited-angle, and truncated projections. This section describes dataset preparation, experimental settings, PEDB implementation details, comparison methods, and evaluation metrics.

\subsection{Dataset Preparation}
\label{sec: dataset}
For simulated experiments, we used the RPLHR-CT-tiny dataset~\citep{yu2022rplhr} and the Mayo Grand Challenge dataset~\citep{Mayo_challenge}. The RPLHR-CT-tiny dataset consists of 50 anonymized chest CT volumes, of which 40 cases (11,090 slices) were used for training and the remaining 10 cases (2,960 slices) were reserved for evaluation. The Mayo dataset comprises anonymized abdominal CT scans from 10 patients, from which 2,900 pelvic slices were selected and used exclusively for evaluation. For both datasets, the original 512 × 512 images were used as ground truth. Complete projections were simulated from these images in a fan-beam geometry with 720 uniformly spaced views over 360 degrees and 800 detector pixels. Detailed geometry settings are provided in \ref{sec: projection geometry}. From the complete projections, we extracted three types of incomplete data: (1) sparse-view projections with 120 uniformly spaced views over 360 degrees; (2) limited-angle projections with 240 contiguous views covering a 120-degree arc; and (3) truncated projections using 400 detector pixels, corresponding to a circular FOV with a diameter equal to approximately 70\% of the width of the square reconstruction region.

For real-data experiments, we employed a cone-beam CT system equipped with a Comet MXR-160HP/11 FB X-ray source and an X-TILE MDBB-ST32 detector comprising 848 lateral × 256 axial pixels. Scans were performed on the Multipurpose Chest Phantom N1 ``LUNGMAN" at 7 uniformly spaced longitudinal bed positions. In each scan, complete projections were acquired from 720 uniformly distributed viewing angles over 360 degrees. To approximate fan-beam geometry, only the 12 central detector rows in the axial direction were used, resulting in a total of 84 sets of complete fan-beam projection data. The corresponding geometry settings are provided in \ref{sec: projection geometry}. All data were used exclusively for evaluation. Ground truth CT images were reconstructed from the full projection data using the FBP algorithm, with an image size of 512 × 512. From the complete projection data, we extracted three types of incomplete projections: (1) sparse-view projections, using 120 uniformly spaced views over 360 degrees; (2) limited-angle projections, using 240 contiguous views spanning a 120-degree arc, consistent with the angular coverage used in the simulated data; and (3) truncated projections, using 384 lateral detector pixels from the full set of 848, corresponding to a circular FOV with a diameter equal to approximately 80\% of the width of the square reconstruction region.

The corrupted images were reconstructed from incomplete projection data using the FBP algorithm. Before that, the incomplete projections were preprocessed according to the type of incompleteness. For sparse-view projections, no preprocessing was applied. For limited-angle projections, Riess reweighting~\citep{riess2013tv} was used to compensate for intensity loss in regions affected by missing rays. For truncated projections, linear extrapolation along the lateral direction was applied to suppress bright ring artifacts near the FOV boundary. These preprocessing approaches were used solely for reconstructing corrupted images; the original, unprocessed projection data was retained for all subsequent processing.

\subsection{Experimental Settings}
\label{sec: experiments}
We conducted extensive experiments to evaluate PEDB in CT reconstruction from three types of incomplete data, including sparse-view, limited-angle, and truncated projections. For each incomplete data type, PEDB was trained separately on paired clean and FBP chest CT images, where the FBP images were reconstructed from simulated incomplete projection data without added noise. For each type, PEDB was then evaluated under four distinct scenarios:\\
(1) Standard simulations.
As a baseline evaluation, PEDB was tested on incomplete projections simulated without added noise using the testing chest CT data. No domain shift was introduced in this scenario.\\
(2) Low-noise and high-noise simulations.
To assess PEDB's performance at noise levels higher than that in training, PEDB without retraining was evaluated on noisy incomplete projections simulated from the same testing chest CT data. Additional quantum noise was introduced using a first-order approximation:
\begin{equation}
\tilde{p}=p+\frac{1}{\sqrt{N_{\text{air}}}}\exp{\left(\frac{p}{2}\right)}z,
\end{equation}
where $z\sim \mathcal{N}\left(0,I\right)$ is Gaussian noise, and $p$ and $\tilde{p}$ represent the original and noise-added post-log projections, respectively. The parameter $N_{\text{air}}$ denotes the number of photons per ray reaching the detector in an air-only path, controlling the noise level. We set $N_{\text{air}} = 2.5 \times 10^5$ for low-noise simulations and $N_{\text{air}} = 4 \times 10^4$ for high-noise simulations. No domain shift was introduced in this scenario.\\
(3) Domain-shift simulations. To evaluate PEDB's generalization ability to unseen anatomical structures, PEDB without retraining was tested on incomplete projections simulated from pelvic CT slices. No additional noise was introduced in this scenario.\\
(4) Real experiments. To assess PEDB's generalization ability under real-world conditions, PEDB without retraining was evaluated on real incomplete projection data. The domain shift in this scenario mainly arises from (i) differences in anatomical structures, as PEDB was trained on human chest CT data but tested on a chest phantom, and (ii) variations in FBP artifact patterns, mainly due to differences in projection geometry between real and simulated data, as summarized in \ref{sec: projection geometry}.

\subsection{Implementation Details for PEDB}
\label{sec: evaluation}

For each type of incomplete data, including sparse-view, limited-angle, and truncated projections, we trained two image-domain data predictors. Within the unified DDBM framework presented in Subsection~\ref{sec: image domain diffusion bridge}, one predictor followed the I$^2$SB specification of the framework’s design space, and the other followed the DDBM specification. The design space includes (1) the formulations of the diffusion coefficient $g(t)$ and the loss weighting function $\lambda(t)$, (2) the network parameterization strategy that expresses the image-domain data predictor $D_{\theta}(X_t, t, X_{\text{FBP}})$ as a linear function of a neural network $F_{\theta}$, and (3) the network preconditioning strategy for rescaling the inputs and outputs of $F_{\theta}$. The specific design choices of the two predictors in these aspects are provided in \ref{sec: specification}.  

Across all three types of incomplete data and both predictor variants, we used the same network architecture and training configuration. The neural network $F_{\theta}$ was implemented as a 2D residual U-Net, following the architecture used in denoising diffusion probabilistic model (DDPM)~\citep{ho2020denoising}. Positional encoding and $X_{\text{FBP}}$ were concatenated with $X_t$ along the channel dimension to serve as conditions for the network. Training was conducted on full 512 × 512 images with a batch size of 32, using the Adam optimizer with a learning rate of $8\times10^{-5}$ for 50000 iterations.

During inference, we used neural function evaluations (NFE) of 10 and 50, with a uniformly spaced time schedule $\{t_i\}_{i=0}^{N}$. The noise scaling factor $\eta_t$ was set to its maximum value, $\eta_{t,\text{max}} = \frac{\sigma_{t-\Delta t}\overline{\sigma}_{t-\Delta t}}{\sigma_T}$, which corresponds to $\gamma \geq 8$. When $\gamma \geq 8$, the relative difference between $\eta_t$ and $\eta_{t,\text{max}}$ is less than 0.1\%. The number of CG iterations $m$ was set to 3 for high-noise simulations, 5 for low-noise simulations, and 20 for all other experiments. The only hyperparameter tuned across different types of incomplete data, different evaluation scenarios, and both predictor variants was $k_x$, which balances the data consistency and regularization terms when incorporating data consistency. The specific values of $k_x$ are provided in Table~\ref{tab: parameters}.  In real experiments, the image-domain expected mean was used as the final output in the last reverse step without incorporating data consistency. In all other scenarios, data consistency was incorporated at each reverse step.
\subsection{Comparison Methods and Evaluation Metrics}
PEDB was compared with methods categorized into five groups: (1) a conventional iterative reconstruction method, TV~\citep{goldstein2009split}; (2) an image-domain end-to-end mapping network, U-Net~\citep{ronneberger2015u}; (3) an unsupervised diffusion model with data consistency, DPS~\citep{chung2022diffusion}, initialized using the jump start technique~\citep{meng2021sdedit}; (4) image domain diffusion bridge models, DDBM~\citep{zhou2023denoising} and I$^2$SB~\citep{liu20232}; and (5) a diffusion bridge model with data consistency, CDDB~\citep{chung2024direct}.

PEDB and the supervised comparison methods, including U-Net, DDBM, I$^2$SB, and CDDB, were trained on paired clean and FBP chest CT images, where the FBP images were reconstructed from simulated incomplete projection data without added noise. The unsupervised method DPS was trained on clean chest CT images. All methods were directly evaluated without retraining across all scenarios described in Subsection~\ref{sec: experiments}.

I$^2$SB and CDDB used the same I$^2$SB-specified image-domain data predictors as PEDB, with NFE set to 10 and 50. DDBM used the same DDBM-specified predictors as PEDB, with NFE set to 299, as it does not achieve satisfactory performance with small NFE values. Inference hyperparameters for all comparison methods were carefully tuned to achieve optimal performance. Additional implementation details for the comparison methods are provided in \ref{sec: comparison_method}.

For quantitative evaluation, we computed Root Mean Square Error (RMSE) and Structural Similarity Index Measure (SSIM) to assess reconstruction fidelity with respect to the ground truth, and LPIPS~\citep{zhang2018perceptual} to evaluate texture restoration quality. 

\section{Results}
This section presents results for PEDB in CT reconstruction from three types of incomplete data, including sparse-view, limited-angle, and truncated projections. Results are presented separately for each incomplete data type, followed by a summary highlighting the overall trends.
\begin{table}[t]
\footnotesize
\begin{center}
\setlength{\tabcolsep}{0.67mm}{
\begin{tabular}{lcccccccccc}
\toprule
Sparse-View&&\multicolumn{3}{c}{Standard}&\multicolumn{3}{c}{Low-Noise}&\multicolumn{3}{c}{High-Noise} \\
\cmidrule(r){3-5}\cmidrule(r){6-8}\cmidrule(r){9-11}
      Method&NFE&RMSE&SSIM&LPIPS&RMSE  &SSIM        &LPIPS &RMSE &SSIM &LPIPS\\
\midrule
TV& 0  & 36.8        & 0.919       &0.299& 41.3        & 0.902        &0.304 & 52.9         & 0.858        &0.353 \\
U-Net& 1  & 29.7     & 0.938        & 0.245     & 31.9      & \textbf{0.929}        & 0.222  &\textbf{41.3}          & \textbf{0.894}        & 0.223\\
DPS&50   & 50.3         &0.866       & 0.191& 50.7         &0.865       & 0.191&52.4            & 0.858         & 0.194         \\
DDBM & 299& 42.1&0.878&0.140&44.9&0.866&0.148&54.8&0.824&0.186\\
I$^2$SB& 10&30.3&0.934&0.218&33.0&0.924&0.200&43.1&0.885&0.203\\
I$^2$SB& 50& 32.8&0.924&0.178 & 35.9&0.912&0.172&46.6&0.870&0.189\\
CDDB& 10&30.1&0.935&0.217 &32.8&0.925&0.199&43.1&0.885&0.201\\
CDDB& 50&32.4&0.926&0.177 &35.5&0.914&0.170&46.3&0.870&0.188\\
\midrule
PEDB (DDBM)$^\dagger$&10&\underline{26.7}&\underline{0.944}&0.138&\underline{31.7}&0.925&0.147&43.2&0.884&0.193\\
PEDB (DDBM)$^\dagger$ &50&27.7&0.940&\textbf{0.118}&33.3&0.917&\textbf{0.129}&45.4&0.872&\textbf{0.176}\\
PEDB (I$^2$SB)$^{\dagger\dagger}$ &10&\textbf{26.4}&\textbf{0.946}&0.143&\textbf{31.3}&\underline{0.926}&0.151&\underline{42.6}&\underline{0.886}&0.196\\
PEDB (I$^2$SB)$^{\dagger\dagger}$ &50&27.0&0.943&\underline{0.131}&32.4&0.921&\underline{0.140}&44.1&0.878&\underline{0.185}\\

\bottomrule
\end{tabular}}
\caption{
Quantitative results for standard and noisy simulations in CT reconstruction from sparse-view projections. Lower RMSE and LPIPS indicate better performance, and higher SSIM is preferred. RMSE is reported in HU. $^{\dagger}$ PEDB with the DDBM-specified image-domain data predictor. $^{\dagger\dagger}$ PEDB with the I$^2$SB-specified image-domain data predictor. \textbf{Bold} indicates the best result; \underline{underline} indicates the second-best.}
\label{tab: sparse_noise}
\end{center}
\end{table}

\begin{table}[t]
\footnotesize
\begin{center}
\setlength{\tabcolsep}{2.52mm}{
\begin{tabular}{lccccccc}
\toprule
 Sparse-View&&\multicolumn{3}{c}{Domain-Shift}&\multicolumn{3}{c}{Real} \\
\cmidrule(r){3-5}\cmidrule(r){6-8}
      Method&NFE&RMSE &SSIM        &LPIPS &RMSE &SSIM        &LPIPS \\
\midrule
TV& 0  & 24.2        & 0.960        &0.227 & 66.3        & 0.858       &0.370 \\
U-Net& 1       & 24.2      & 0.960       & 0.198 & 47.3         & 0.906        & 0.242\\
DPS&$\geq20^{\dagger\dagger\dagger}$  & 49.8     &0.848      & 0.334&77.6            & 0.813        & 0.292         \\
DDBM & 299&44.4&0.857&0.280&55.4&0.863&0.226\\
I$^2$SB& 10&24.8&0.957&0.162&45.4&0.903&0.231\\
I$^2$SB& 50 & 28.6&0.941&0.176&47.7&0.896&0.222\\
CDDB& 10 &24.4&0.958&0.160&45.0&0.906&0.227\\
CDDB& 50 &28.0&0.943&0.173&47.1&0.900&0.217\\
\midrule
PEDB (DDBM)$^\dagger$ &10&\underline{21.2}&\underline{0.963}&0.135&\underline{43.9}&\textbf{0.913}&0.220\\
PEDB (DDBM)$^\dagger$ &50&23.7&0.953&0.152&44.8&0.901&\textbf{0.210}\\
PEDB (I$^2$SB)$^{\dagger\dagger}$ &10&\textbf{19.8}&\textbf{0.968}&\textbf{0.128}&\textbf{43.7}&\underline{0.911}&0.223\\
PEDB (I$^2$SB)$^{\dagger\dagger}$ &50&\underline{21.2}&0.962&\underline{0.133}&\underline{43.9}&0.907&\underline{0.211}\\
\bottomrule
\end{tabular}}
\caption{
Quantitative results for domain-shift simulations and real experiments in CT reconstruction from sparse-view projections. Lower RMSE and LPIPS indicate better performance, and higher SSIM is preferred. RMSE is reported in HU. $^{\dagger}$ PEDB with the DDBM-specified image-domain data predictor. $^{\dagger\dagger}$ PEDB with the I$^2$SB-specified image-domain data predictor. $^{\dagger\dagger\dagger}$ For DPS, NFE is set to 20 in domain-shift simulations and 50 in real experiments. \textbf{Bold} indicates the best result; \underline{underline} indicates the second-best.}
\label{tab: sparse_domain_shift}
\end{center}
\end{table}
\begin{figure*}[!t]
\centering
\includegraphics[width=\textwidth]{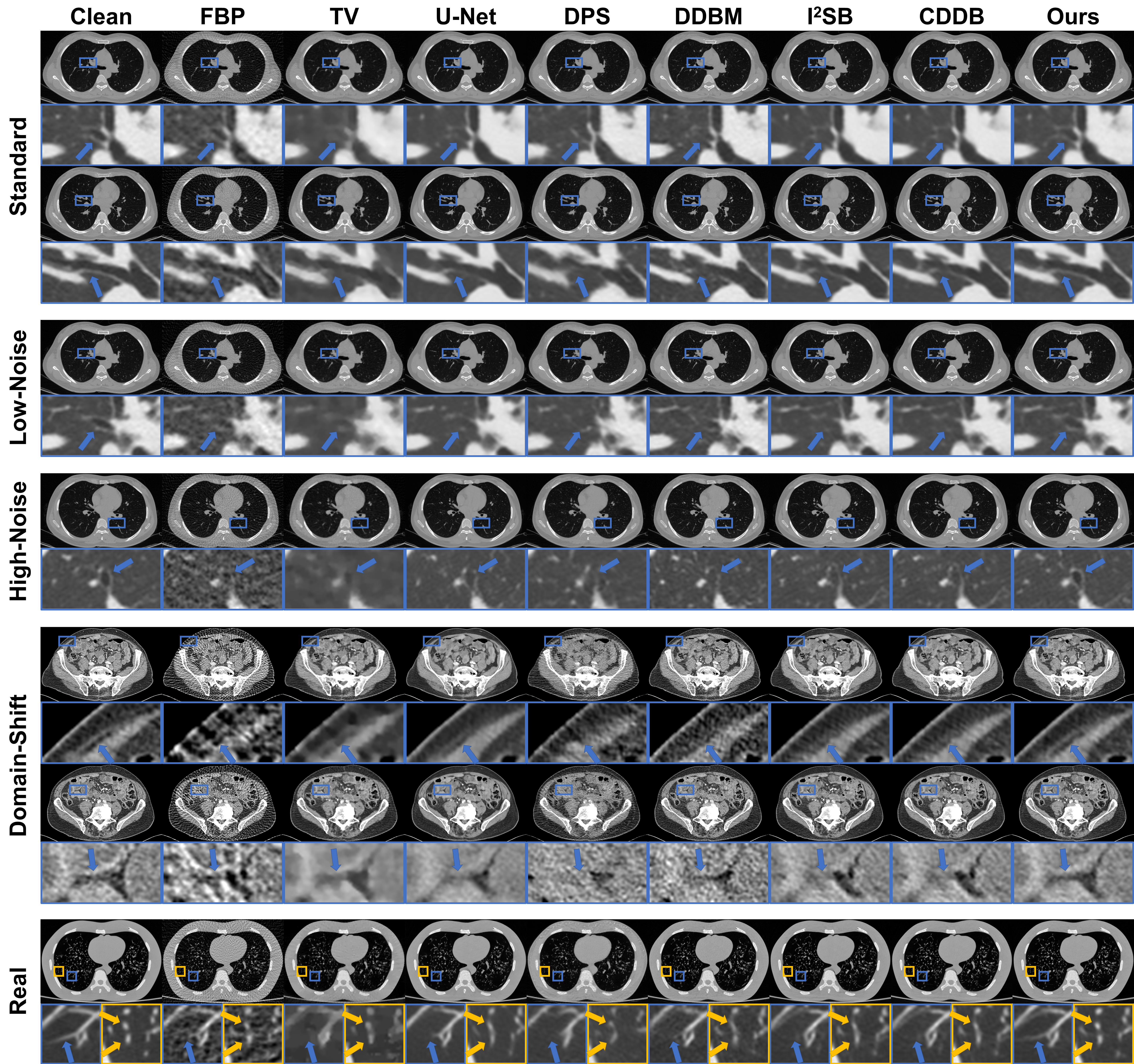}
\caption{Representative visualization results for CT reconstruction from sparse-view projections. The regions enclosed by blue and yellow boxes are magnified to highlight structural details. The display window is set to [-1000HU, 800HU] for full chest CT images, 
[-160HU, 240HU] for full pelvic CT images and zoomed-in soft tissue regions, and [-1300HU, 200HU] for zoomed-in lung regions. Our method PEDB uses the I$^2$SB-specified image-domain data predictor with 50 NFEs. I$^2$SB and CDDB are also evaluated with 50 NFEs.}
\label{fig: sparse}
\end{figure*}
\subsection{Results for CT Reconstruction from Sparse-View Projections}
We present quantitative results for sparse-view CT reconstruction in Tables~\ref{tab: sparse_noise} and~\ref{tab: sparse_domain_shift}. Across all evaluation scenarios, PEDB outperformed all other evaluated diffusion bridge models. Compared with DDBM, PEDB using the same image-domain data predictor (PEDB (DDBM)) at NFE = 50 achieved substantial RMSE reductions (17\% to 47\%) along with higher SSIM and lower LPIPS, demonstrating improved reconstruction fidelity. Compared with I$^2$SB and CDDB, PEDB using the same image-domain data predictor (PEDB (I$^2$SB)) achieved superior results across all three metrics when evaluated at each matched NFE (10 and 50). In addition, PEDB outperformed TV and DPS under all evaluation scenarios and surpassed U-Net across standard simulations, domain-shift simulations and real experiments.

When compared with U-Net in noisy scenarios for sparse-view CT reconstruction (Table \ref{tab: sparse_noise}), PEDB with the I$^2$SB-specified predictor (PEDB (I$^2$SB)) at NFE = 10 achieved a 32\% reduction in LPIPS in the low-noise simulations while maintaining comparable RMSE and SSIM. In the high-noise simulations, it still achieved a 12\% reduction in LPIPS, albeit with slightly worse RMSE and SSIM. U-Net, which can be interpreted as a single-step image-domain diffusion bridge (NFE = 1), tends to produce smoother and less distorted images due to the regression-to-the-mean effect~\citep{delbracio2023inversion}. This leads to lower RMSE and higher SSIM but comes at the cost of reduced perceptual quality. In contrast, our method with a larger NFE is capable of recovering richer textures and finer details, as indicated by significantly lower LPIPS. At the same time, this improvement in perceptual quality may introduce additional deviations from the ground truth during multi-step sampling~\citep{chung2024direct,wang2025implicit}. In the high-noise simulations, such distortions may not be fully corrected by data consistency, resulting in slight compromises in RMSE and SSIM.

The representative visualization results for sparse-view CT reconstruction are shown in Figure~\ref{fig: sparse}. Compared to all other methods, PEDB achieved more accurate detail restoration, particularly in anatomical regions such as the pulmonary airways in both standard and noisy simulations, subcutaneous adipose tissue in domain-shift simulations, and pulmonary vessels in real experiments. Moreover, PEDB demonstrated strong generalization ability to unseen anatomical structures by successfully recovering adipose tissue in the inter-intestinal spaces in the domain-shift simulations, despite this structure being absent from the training data.

\begin{table}[t]
\footnotesize
\begin{center}
\setlength{\tabcolsep}{0.67mm}{
\begin{tabular}{lcccccccccc}
\toprule
 Limited-Angle&&\multicolumn{3}{c}{Standard}&\multicolumn{3}{c}{Low-Noise}&\multicolumn{3}{c}{High-Noise} \\
\cmidrule(r){3-5}\cmidrule(r){6-8}\cmidrule(r){9-11}
      Method&NFE&RMSE&SSIM&LPIPS&RMSE  &SSIM        &LPIPS &RMSE &SSIM &LPIPS\\
\midrule
TV& 0  & 105.5       & 0.862        &0.265&105.5        & 0.856       &0.291 & 109.2        & \underline{0.813}       &0.314 \\
U-Net& 1  & 36.4    & 0.954       & 0.157     & 42.8      & 0.923        & 0.176  &66.8          &0.796       & 0.256\\
DPS&600  & 70.7        &0.842       & 0.204& 70.7         &0.843     & 0.204&70.8           & \textbf{0.844}        & \textbf{0.206}        \\
DDBM & 299&41.5&0.920&\underline{0.110}&48.4&0.887&\textbf{0.143}&74.5&0.755&\underline{0.243}\\
I$^2$SB& 10&34.4&0.950&0.153&41.1&0.921&0.174&64.9&0.804&0.254\\
I$^2$SB& 50&35.9&0.945&0.139 & 42.3&0.917&0.167&65.7&0.802&0.253\\
CDDB& 10&33.2&0.956&0.152&40.1&0.925&0.174&63.9&0.809&0.254\\
CDDB& 50 &34.5&0.952&0.138 &41.2&0.922&0.167&64.5&0.806&0.252\\
\midrule
PEDB (DDBM)$^\dagger$ &10&29.2&0.962&0.125&38.8&0.925&0.164&64.5&0.803&0.250\\
PEDB (DDBM)$^\dagger$ &50&\underline{29.0}&0.960&\textbf{0.101}&38.9&0.921&\underline{0.148}&63.6&0.800&\underline{0.243}\\
PEDB (I$^2$SB)$^{\dagger\dagger}$ &10&29.4&\textbf{0.963}&0.133&\underline{38.3}&\textbf{0.927}&0.172&\underline{62.3}&0.809&0.253\\
PEDB (I$^2$SB)$^{\dagger\dagger}$ &50&\textbf{28.4}&\textbf{0.963}&0.123&\textbf{37.7}&\underline{0.926}&0.165&\textbf{61.7}&0.809&0.251\\

\bottomrule
\end{tabular}}
\caption{
Quantitative results for standard and noisy simulations in CT reconstruction from limited-angle projections. Lower RMSE and LPIPS indicate better performance, and higher SSIM is preferred. RMSE is reported in HU. $^{\dagger}$ PEDB with the DDBM-specified image-domain data predictor. $^{\dagger\dagger}$ PEDB with the I$^2$SB-specified image-domain data predictor. \textbf{Bold} indicates the best result; \underline{underline} indicates the second-best.}
\label{tab: limit_noise}
\end{center}
\end{table}

\begin{table}[t]
\footnotesize
\begin{center}
\setlength{\tabcolsep}{2.77mm}{
\begin{tabular}{lccccccc}
\toprule
 Limited-Angle&&\multicolumn{3}{c}{Domain-Shift}&\multicolumn{3}{c}{Real} \\
\cmidrule(r){3-5}\cmidrule(r){6-8}
      Method&NFE&RMSE &SSIM        &LPIPS &RMSE &SSIM        &LPIPS \\
\midrule
TV& 0  & 72.3
& 0.927        &0.225 & 167.9        & 0.762       &0.405 \\
U-Net& 1       & 51.7      & 0.933       & 0.180 & 105.9         & 0.833        & 0.203\\
DPS&400  & 77.3     &0.815      & 0.366&177.5            & 0.697        & 0.351         \\
DDBM & 299&58.0&0.879&0.264&114.0&0.800&\textbf{0.191}\\
I$^2$SB& 10&48.8&0.935&0.158&110.6&0.829&0.219\\
I$^2$SB& 50 & 50.8&0.930&0.156&111.3&0.827&0.215\\
CDDB& 10 &43.8&0.957&0.156&102.3&0.844&0.217\\
CDDB& 50 &45.5&0.951&0.153&101.8&0.845&0.213\\
\midrule
PEDB (DDBM)$^\dagger$ &10&38.1&0.964&\underline{0.126}&92.9&0.872&\underline{0.196}\\
PEDB (DDBM)$^\dagger$ &50&\underline{36.7}&0.959&0.137&94.2&0.860&\underline{0.196}\\
PEDB (I$^2$SB)$^{\dagger\dagger}$ &10&37.9&\textbf{0.967}&0.135&\underline{92.0}&\underline{0.873}&0.210\\
PEDB (I$^2$SB)$^{\dagger\dagger}$ &50&\textbf{36.5}&\textbf{0.967}&\textbf{0.123}&\textbf{84.5}&\textbf{0.877}&0.203\\
\bottomrule
\end{tabular}}
\caption{
Quantitative results for domain-shift simulations and real experiments in CT reconstruction from limited-angle projections. Lower RMSE and LPIPS indicate better performance, and higher SSIM is preferred. RMSE is reported in HU. $^{\dagger}$ PEDB with the DDBM-specified image-domain data predictor. $^{\dagger\dagger}$ PEDB with the I$^2$SB-specified image-domain data predictor. \textbf{Bold} indicates the best result; \underline{underline} indicates the second-best.}
\label{tab: limit_domain_shift}
\end{center}
\end{table}
\begin{figure*}[!t]
\centering
\includegraphics[width=\textwidth]{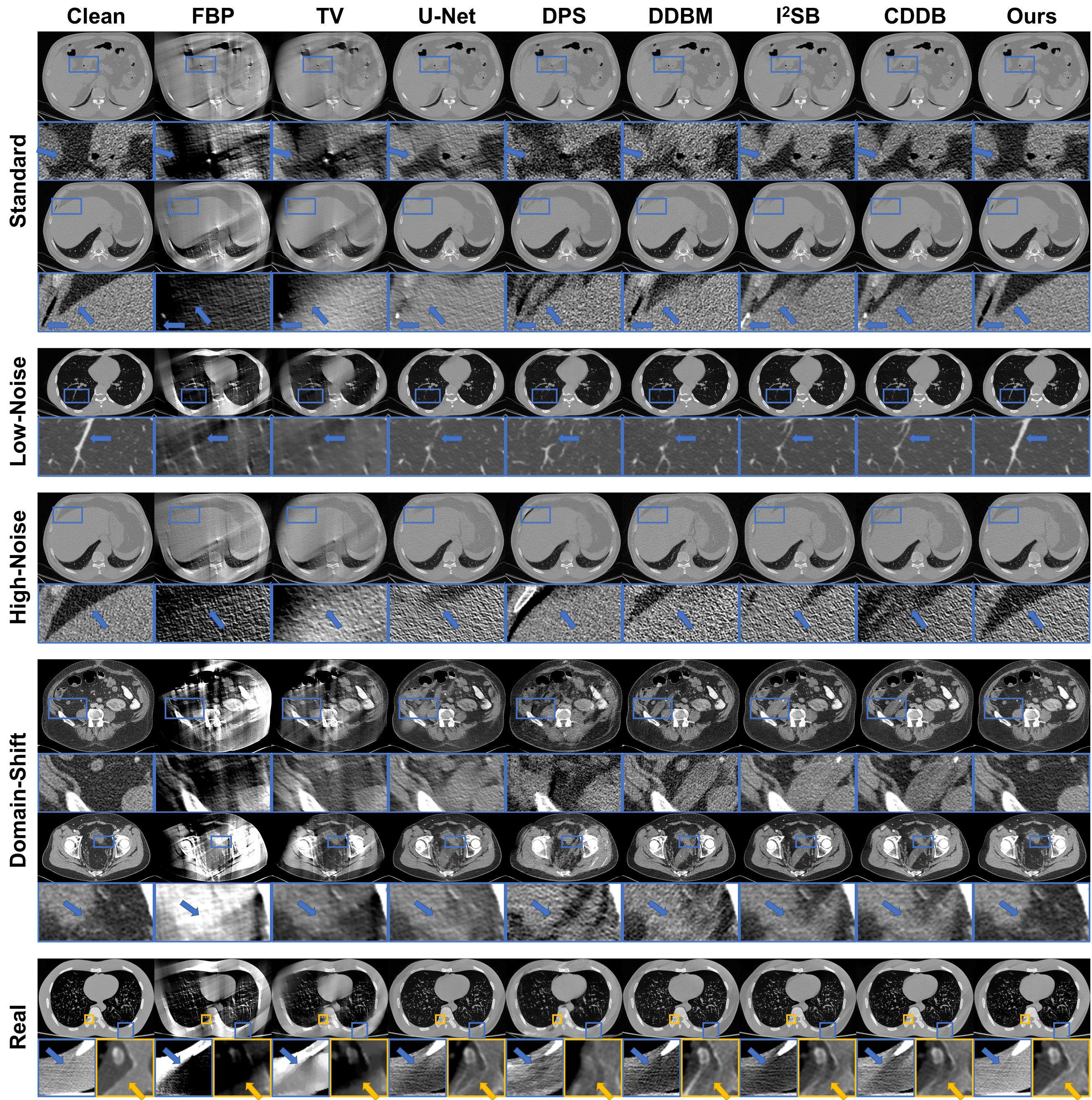}
\caption{Representative visualization results for CT reconstruction from limited-angle projections. The regions enclosed by blue and yellow boxes are magnified to highlight structural details. The display window is set to [-1000HU, 800HU] for full chest CT images, 
[-160HU, 240HU] for full pelvic CT images and zoomed-in soft tissue regions, [-1300HU, 200HU] for zoomed-in lung regions, and [-500HU, 1500HU] for zoomed-in bone regions in real experiments. Our method PEDB uses the I$^2$SB-specified image-domain data predictor with 50 NFEs. I$^2$SB and CDDB are also evaluated with 50 NFEs.}
\label{fig: limit}
\end{figure*}
\subsection{Results for CT Reconstruction from Limited-Angle Projections}
We present quantitative results for limited-angle CT reconstruction in Tables~\ref{tab: limit_noise} and~\ref{tab: limit_domain_shift}. Across all evaluation scenarios, PEDB outperformed all other evaluated diffusion bridge models. Compared with DDBM, PEDB using the same image-domain data predictor (PEDB (DDBM)) at NFE = 50 achieved substantial RMSE reductions (15\% to 37\%) along with higher SSIM and comparable LPIPS, demonstrating improved reconstruction fidelity. Compared with I$^2$SB and CDDB, PEDB using the same image-domain data predictor (PEDB (I$^2$SB)) achieved superior results across all three metrics when evaluated at each matched NFE (10 and 50). In addition, PEDB outperformed TV and U-Net under all evaluation scenarios and surpassed DPS across standard, low-noise, and domain-shift simulations, as well as real experiments.

When compared with DPS in the high-noise simulations for limited-angle CT reconstruction (Table \ref{tab: limit_noise}), PEDB achieved better RMSE while exhibiting worse SSIM and LPIPS. From the visualization results, we observe that, compared to DPS, PEDB tends to produce reconstructions with higher noise levels, leading to worse SSIM and LPIPS, but it generally introduces fewer hallucinations and recovers anatomical structures with higher fidelity, which corresponds to better RMSE. An example is shown in the "High-Noise" row of Figure~\ref{fig: limit}.  In the zoomed-in region enclosed by the blue box, DPS produced an image with a noise level similar to the ground truth but incorrectly generated rib-like structures with lung-air space in the region that should contain visceral adipose tissue around the liver, as seen in the clean image. In contrast, our method correctly restored the anatomical structure of the adipose tissue, albeit with a higher noise level in the reconstructed image. Additional visualization results comparing DPS and PEDB in this scenario are provided in \ref{sec: additional_visual}. The high noise level in PEDB's reconstruction mainly arises from the noisy FBP reconstruction. Since the image-domain data predictor was trained on simulated data without added noise, noisy FBP reconstruction results in the predicted image-domain expected mean being noisy. Incorporating data consistency through equation~(\ref{eq:optimization}) is unable to suppress this noise.

The representative visualization results for limited-angle CT reconstruction are shown in Figure~\ref{fig: limit}. Compared to all other methods, PEDB achieved more accurate detail restoration, particularly in anatomical regions such as the adipose tissue surrounding the liver in standard and high-noise simulations, pulmonary vessels in low-noise simulations, and the borders of the spinal vertebrae in real experiments. Notably, PEDB was the only method that fully corrected the limited-angle artifact within the posterior thoracic region enclosed by the blue box in the real data, demonstrating its strong generalization ability to geometry-induced variations in FBP artifact patterns. Furthermore, PEDB also demonstrated strong generalization ability to unseen anatomical structures by successfully recovering adipose tissue adjacent to the pelvic bones and within the urogenital region in domain-shift simulations, despite these structures being absent from the training data.
\begin{table}[t]
\footnotesize
\begin{center}
\setlength{\tabcolsep}{0.67mm}{
\begin{tabular}{lcccccccccc}
\toprule
 Truncated&&\multicolumn{3}{c}{Standard}&\multicolumn{3}{c}{Low-Noise}&\multicolumn{3}{c}{High-Noise} \\
\cmidrule(r){3-5}\cmidrule(r){6-8}\cmidrule(r){9-11}
      Method&NFE&RMSE&SSIM&LPIPS&RMSE  &SSIM        &LPIPS &RMSE &SSIM &LPIPS\\
\midrule
TV& 0  & 175.3       & 0.794        &0.246&174.4        & 0.783       &0.280 & 174.1        & 0.757      &0.361 \\
U-Net& 1  & 118.0    & 0.910       & 0.148     & 119.3     & 0.888        & 0.169  &125.7          &0.801       & 0.240\\
DPS&300  & 150.9        &0.761       & 0.250& 150.8         &0.762    & 0.250&150.3           & 0.765        & 0.250        \\
DDBM & 299&61.6&0.923&0.100&71.8&0.896&0.126&103.6&0.787&0.208\\
I$^2$SB& 10&51.2&0.956&0.104&60.2&0.931&0.124&91.4&0.827&0.202\\
I$^2$SB& 50&52.7&0.953&0.095&61.3&0.927&0.118&92.0&0.824&\underline{0.199}\\
CDDB& 10&50.5&0.957&0.104&59.4&0.932&0.124&90.5&\underline{0.828}&0.201\\
CDDB& 50 &51.8&0.953&0.095 &60.4&0.928&0.118&90.9&0.825&\underline{0.199}\\
\midrule
PEDB (DDBM)$^\dagger$ &10&47.2&0.966&0.076&60.7&0.929&0.123&94.4&0.823&0.206\\
PEDB (DDBM)$^\dagger$ &50&47.2&0.968&\textbf{0.059}&57.5&0.929&\textbf{0.111}&\underline{86.5}&0.822&0.200\\
PEDB (I$^2$SB)$^{\dagger\dagger}$ &10&\underline{44.3}&\underline{0.969}&0.076&\underline{55.7}&\textbf{0.934}&0.121&88.3&\underline{0.828}&0.200\\
PEDB (I$^2$SB)$^{\dagger\dagger}$ &50&\textbf{43.9}&\textbf{0.970}&\underline{0.067}&\textbf{54.5}&\textbf{0.934}&\underline{0.114}&\textbf{84.2}&\textbf{0.830}&\textbf{0.196}\\
\bottomrule
\end{tabular}}
\caption{
Quantitative results for standard and noisy simulations in CT reconstruction from truncated projections. Lower RMSE and LPIPS indicate better performance, and higher SSIM is preferred. RMSE is reported in HU. $^{\dagger}$ PEDB with the DDBM-specified image-domain data predictor. $^{\dagger\dagger}$ PEDB with the I$^2$SB-specified image-domain data predictor. \textbf{Bold} indicates the best result; \underline{underline} indicates the second-best.}
\label{tab:trun_noise}
\end{center}
\end{table}

\begin{table}[t]
\footnotesize
\begin{center}
\setlength{\tabcolsep}{2.41mm}{
\begin{tabular}{lccccccc}
\toprule
 Truncated&&\multicolumn{3}{c}{Domain-Shift}&\multicolumn{3}{c}{Real} \\
\cmidrule(r){3-5}\cmidrule(r){6-8}
      Method&NFE&RMSE &SSIM        &LPIPS &RMSE &SSIM        &LPIPS \\
\midrule
TV& 0  & 155.5
& 0.828        &0.233 & 209.6        & 0.758       &0.364 \\
U-Net& 1       & 153.2     & 0.870       & 0.187 & 246.9         & 0.761        & 0.211\\
DPS&$\geq300^{\dagger\dagger\dagger}$  & 153.2    &0.740      & 0.402&208.4            & 0.676        & 0.350         \\
DDBM & 299&130.8&0.858&0.266&225.2&0.767&0.214\\
I$^2$SB& 10&132.0&0.898&0.147&233.6&0.733&0.203\\
I$^2$SB& 50 & 138.4&0.893&0.159&240.2&0.731&0.210\\
CDDB& 10 &130.9&0.902&0.147&202.3&0.762&0.191\\
CDDB& 50 &136.8&0.897&0.159&204.5&0.772&0.195\\
\midrule
PEDB (DDBM)$^\dagger$ &10&\textbf{123.4}&0.909&0.132&\underline{180.4}&0.838&\textbf{0.170}\\
PEDB (DDBM)$^\dagger$ &50&\underline{127.8}&\underline{0.912}&0.137&185.2&0.820&0.187\\
PEDB (I$^2$SB)$^{\dagger\dagger}$ &10&130.5&\underline{0.912}&\underline{0.130}&185.8&\underline{0.840}&0.175\\
PEDB (I$^2$SB)$^{\dagger\dagger}$ &50&130.8&\textbf{0.914}&\textbf{0.126}&\textbf{180.2}&\textbf{0.853}&\textbf{0.170}\\
\bottomrule
\end{tabular}}
\caption{
Quantitative results for domain-shift simulations and real experiments in CT reconstruction from truncated projections. Lower RMSE and LPIPS indicate better performance, and higher SSIM is preferred. RMSE is reported in HU. $^{\dagger}$ PEDB with the DDBM-specified image-domain data predictor. $^{\dagger\dagger}$ PEDB with the I$^2$SB-specified image-domain data predictor. $^{\dagger\dagger\dagger}$ For DPS, NFE is set to 400 in domain-shift simulations and 300 in real experiments. \textbf{Bold} indicates the best result; \underline{underline} indicates the second-best.}
\label{tab: truncation_domain_shift}
\end{center}
\end{table}
\begin{figure*}[!t]
\centering
\includegraphics[width=\textwidth]{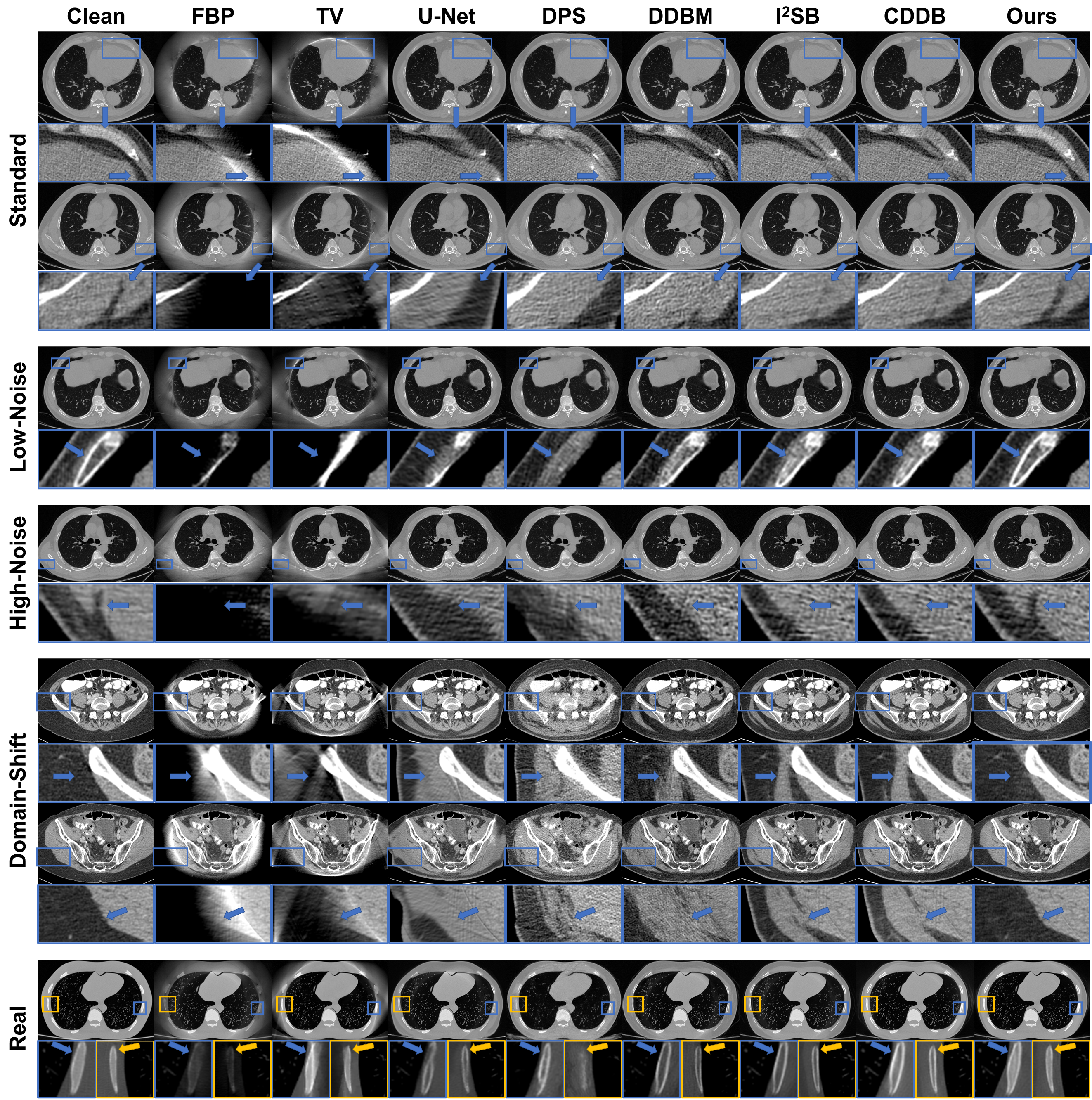}
\caption{Representative visualization results for CT reconstruction from truncated projections. The regions enclosed by blue and yellow boxes are magnified to highlight structural details. The display window is set to [-1000HU, 800HU] for full chest CT images, 
[-160HU, 240HU] for full pelvic CT images and zoomed-in soft tissue regions, and [-500HU, 1500HU] for zoomed-in bone regions in real experiments. Our method PEDB uses the I$^2$SB-specified image-domain data predictor with 50 NFEs. I$^2$SB and CDDB are also evaluated with 50 NFEs.}
\label{fig: truncated}
\end{figure*}
\subsection{Results for CT Reconstruction from Truncated Projections}
We present quantitative results for CT reconstruction from truncated projections in Tables~\ref{tab:trun_noise} and~\ref{tab: truncation_domain_shift}. Across all evaluation scenarios, PEDB achieved the best performance among all comparison methods. Compared with DDBM, PEDB achieved superior results across all three metrics. Compared with I$^2$SB and CDDB, PEDB using the same image-domain data predictor (PEDB (I$^2$SB)) also achieved superior results across all metrics when evaluated at each matched NFE (10 and 50). Notably, in the real experiments (Table~\ref{tab: truncation_domain_shift}), PEDB (I$^2$SB) outperformed I$^2$SB and CDDB by a large margin, achieving an 8\% to 25\% reduction in RMSE, a 10\% to 17\% increase in SSIM, and an 8\% to 19\% reduction in LPIPS, demonstrating superior generalization ability under domain shift. In addition, PEDB outperformed TV, U-Net, and DPS in all metrics across all evaluation scenarios.

The superior performance of PEDB in CT reconstruction from truncated projections is also supported by the representative visualization results in Figure~\ref{fig: truncated}. Compared to all other methods, PEDB achieved more accurate detail restoration, particularly in anatomical regions such as subcutaneous adipose tissue and adipose tissue around the heart in standard and high-noise simulations, as well as the ribs in low-noise simulations. Furthermore, PEDB demonstrated strong generalization ability to unseen anatomical structures by accurately reconstructing thicker subcutaneous adipose tissue in pelvic slices in domain-shift simulations and recovering the solid bone structures of the chest phantom in real experiments, despite these structures being absent from the training data.
\subsection{Summary for Results}
PEDB demonstrated strong performance in CT reconstruction from three types of incomplete data, including sparse-view, limited-angle and truncated projections. For each of these types, PEDB outperformed all other evaluated diffusion bridge models across all evaluation scenarios. The quantitative results demonstrate PEDB's improved reconstruction fidelity and enhanced perceptual quality, and the representative visualization results show that PEDB more effectively recovers anatomical structures, even when such structures are absent from the training data. In addition, PEDB generally outperformed TV, U-Net, and DPS.

\section{Ablation Studies}
This section presents ablation studies for PEDB on two key hyperparameters: $k_x$ and $\gamma$. The parameter $k_x$ balances the data consistency term and the regularization term when incorporating data consistency, affecting PEDB's performance when evaluated on noisy projections. This effect was examined in the low-noise simulation scenario. The parameter $\gamma$ controls the intensity of the Langevin term in the proposed reverse SDE~(\ref{eq: reverse sde}), affecting PEDB's generalization ability under domain shift. This effect was evaluated in the real experiment scenario.
\begin{figure*}[!t]
\centering
\includegraphics[width=\textwidth]{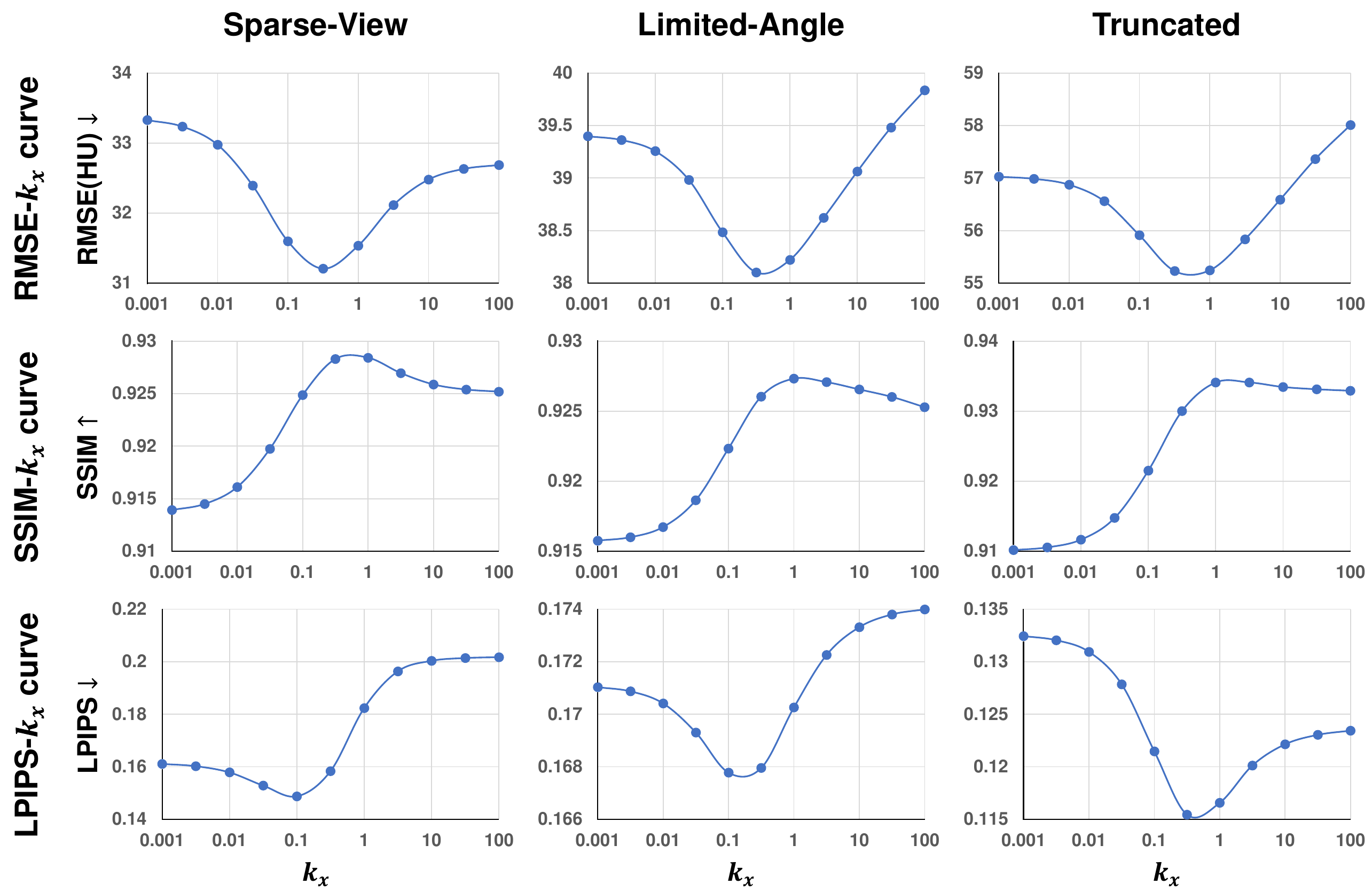}
\caption{Ablation study on the hyperparameter $k_x$, showing RMSE–$k_x$, SSIM–$k_x$, and LPIPS–$k_x$ curves across three types of incomplete data, including sparse-view, limited-angle, and truncated projections. Evaluations were conducted in the low-noise simulation scenario. I$^2$SB-specified image-domain data predictors were used, with NFE set to 10 and the number of CG iterations $m$ set to 20. The hyperparameter $k_x$ was varied from 0.001 to 100.}
\label{fig: k_x}
\end{figure*}
\subsection{Ablation Study on $k_x$}
We investigated the effect of the hyperparameter $k_x$ on PEDB’s performance when evaluated on noisy projections. Evaluations were conducted in the low-noise simulation scenario across three types of incomplete data, including sparse-view, limited-angle, and truncated projections. I$^2$SB-specified image-domain data predictors were used with NFE set to 10, and $k_x$ was varied from 0.001 to 100. To reduce early stopping regularization in the CG method and better isolate the effect of the regularization term in equation~(\ref{eq:optimization}), the number of CG iterations $m$ was increased to 20. All other implementation details followed those used in low-noise simulations described in Subsection~\ref{sec: evaluation}. For quantitative analysis, we plotted RMSE-$k_x$, SSIM-$k_x$, and LPIPS-$k_x$ curves, as shown in Figure~\ref{fig: k_x}.

The curves in Figure~\ref{fig: k_x} exhibit a consistent trend: as $k_x$ increases, the quantitative metrics become better, reach optimal values, and then degrade. This trend arises from $k_x$’s role in balancing the data consistency term and the regularization term when obtaining the projection-embedded expected mean $\hat{X}_{0,\text{p}}^{(t)}$ from the image-domain expected mean $\hat{X}_0^{(t)}$ and the projection data $y$ via equation~(\ref{eq:optimization}). Based on these curves, a suitable range for $k_x$ in this setting is approximately 0.1 to 1. When $k_x$ is too small, the regularization term is insufficient, resulting in $\hat{X}_{0,\text{p}}^{(t)}$ being heavily influenced by measurement noise in $y$. Conversely, when $k_x$ is too large, the regularization term dominates, hindering the incorporation of data consistency.
\begin{figure*}[!t]
\centering
\includegraphics[width=\textwidth]{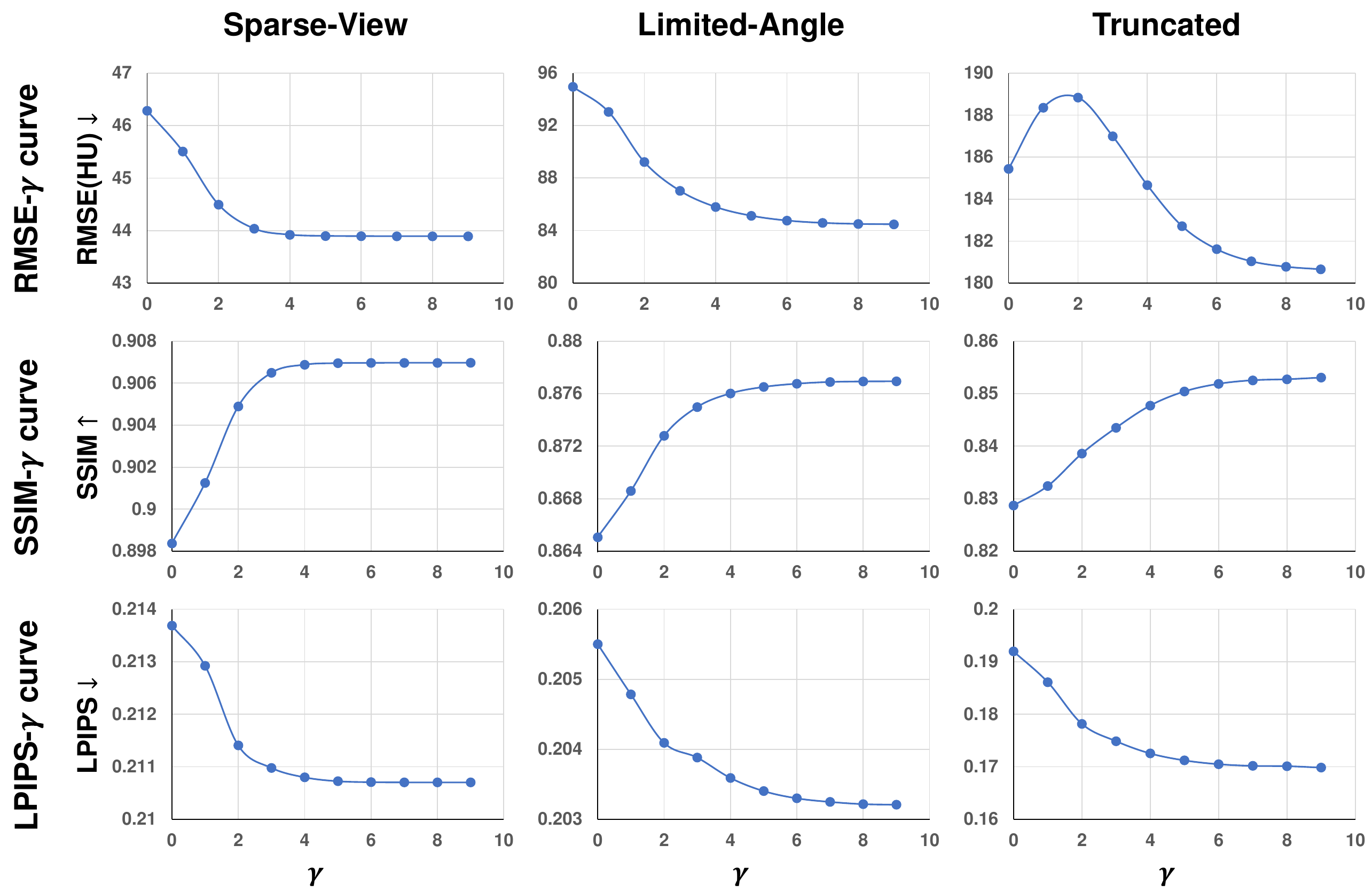}
\caption{Ablation study on the hyperparameter $\gamma$, showing RMSE-$\gamma$, SSIM-$\gamma$, and LPIPS-$\gamma$ curves across three types of incomplete data, including sparse-view, limited-angle, and truncated projections. Evaluations were conducted in the real experiment scenario. I$^2$SB-specified image-domain data predictors were used, with NFE set to 50. The hyperparameter $\gamma$ was varied from 0 to 9.}
\label{fig: gamma}
\end{figure*}
\subsection{Ablation Study on $\gamma$}
\label{sec: ablation_gamma}
We investigated the effect of the hyperparameter $\gamma$ on PEDB’s generalization ability under domain shift. Evaluations were conducted in the real experiment scenario across three types of incomplete data, including sparse-view, limited-angle, and truncated projections. I$^2$SB-specified image-domain data predictors were used with NFE set to 50, and $\gamma$ was varied from 0 to 9. All other implementation details followed those used in real experiments described in Subsection~\ref{sec: evaluation}. For quantitative analysis, we plotted RMSE-$\gamma$, SSIM-$\gamma$, and LPIPS-$\gamma$ curves, as shown in Figure~\ref{fig: gamma}.

The curves in Figure~\ref{fig: gamma} show that PEDB generally achieves better performance with larger values of $\gamma$ in real experiments. This performance gain stems from the increased contribution of the Langevin term in the proposed reverse SDE~(\ref{eq: reverse sde}), which helps pull the intermediate image $X_t$ back toward the target marginal distribution $q\left(X_t|X_{\text{FBP}},y\right)$ when $X_t$ deviates from it due to domain shift, as explained in Subsection~\ref{section: reverse sde}. Additionally, the curves show that the quantitative metrics vary only slightly with respect to $\gamma$ once $\gamma \geq 6$, suggesting that the exact value of $\gamma$ becomes less critical beyond this point. Therefore, instead of carefully tuning $\gamma$, which may be challenging when $k_x$ also needs to be tuned, we can simply set the noise scaling factor $\eta_t$ to $\frac{\sigma_{t-\Delta t}\overline{\sigma}_{t-\Delta t}}{\sigma_T}$, which corresponds to a sufficiently large $\gamma$ according to equation~(\ref{eq: eta_t}).

\section{Discussion}

PEDB is a diffusion bridge model that introduces a novel reverse SDE to sample from the distribution of clean images conditioned on both the FBP reconstruction and the incomplete projection data.  It outperforms all other evaluated diffusion bridge models across all three types of incomplete data and all evaluation scenarios. These performance gains primarily stem from PEDB’s effective integration of data consistency, which helps the reconstructed image align more closely with the observed projection data and enables better utilization of the structural information contained in the projections. In addition, the designed discretization scheme in PEDB also contributes to its superior performance over DDBM. While DDBM directly applies the Euler–Maruyama method to discretize its reverse SDE~(\ref{eq: image domain reverse sde}), PEDB introduces a discretization scheme for the reverse SDE~(\ref{eq: reverse sde}), thereby eliminating discretization errors in both the linear drift and stochastic noise terms.

Compared with the unsupervised diffusion model DPS, PEDB achieves markedly higher reconstruction fidelity, as supported by its substantial RMSE reduction across all three types of incomplete data and all evaluation scenarios. This performance improvement primarily arises from the supervised learning mechanism in PEDB. The image-domain data predictor in PEDB is trained using the FBP reconstruction as a condition, enabling PEDB to better utilize and preserve the structural information contained in the FBP images. In addition, PEDB demonstrates significantly stronger generalization ability to unseen anatomical structures than DPS, as reflected by PEDB's substantial performance gains across all three metrics under domain-shift simulations for all three incomplete data types. The unconditional prior score function in DPS tends to generate details learned from the training data, and these details may not align with anatomical structures unseen during training. In contrast, the supervised image-domain data predictor in PEDB  better focuses on correcting artifacts caused by data incompleteness and produces fewer inconsistent details. Combined with the effective incorporation of data consistency, PEDB thus achieves more faithful recovery of unseen anatomical structures.

Despite these advantages, our approach has certain limitations. First, compared with the unsupervised method DPS, PEDB trained at a specific noise level is more sensitive to noise increases during evaluation. As the noise level increases from standard to low-noise and then to high-noise simulations, PEDB’s quantitative performance degrades, whereas DPS remains relatively stable. PEDB’s noise sensitivity primarily stems from its supervised image-domain data predictor. Because the predictor takes the FBP reconstruction as input, its predicted image-domain expected mean can be adversely affected by noise increases in the FBP image. In contrast, the unconditional prior score used in DPS does not rely on the FBP reconstruction, making DPS inherently more robust to noise. A potential way to mitigate this limitation is to train PEDB on mixed noise levels, which could enhance its robustness across a broader noise range.

Second, PEDB relies on Gaussian assumptions for both the FBP-conditioned clean image distribution and the projection data likelihood. While these assumptions enable a tractable form of the posterior score function and lead to an algorithm that has demonstrated effectiveness through extensive experiments, they may not perfectly capture the true underlying distributions. Refining these assumptions while maintaining tractability of the posterior score represents an important direction for future improvement.

Third, incorporating data consistency in PEDB can be time-consuming, as it requires solving the quadratic optimization problem in equation~(\ref{eq:optimization}) at each reverse step. While we approximate the exact solution using an $m$-step CG update initialized from the image-domain expected mean, this procedure can still be computationally demanding. The inference times for PEDB and the comparison methods are reported in ~\ref{sec: time}. One potential way to alleviate this limitation is to integrate acceleration techniques commonly used in iterative reconstruction, such as Ordered Subsets~\citep{hudson1994accelerated}, which could reduce the computation time required for incorporating data consistency.

In addition to addressing the limitations discussed above, we plan to extend PEDB in two directions. In this work the corrupted image is taken as the FBP reconstruction, but the framework is not restricted to this choice and can accommodate corrupted images obtained by any method. For instance, the corrupted image could be obtained using a U-Net, a diffusion model, or a diffusion bridge model, after which PEDB could be applied to sample from the distribution of clean images conditioned jointly on the corrupted image and the incomplete projection data. Exploring such alternative corrupted image settings may further enhance PEDB’s performance. We also plan to apply PEDB to three-dimensional CT reconstruction tasks, broadening its application beyond two-dimensional scenarios.
\section{Conclusion}
In conclusion, we propose PEDB for CT reconstruction from incomplete data. PEDB introduces a novel reverse SDE to sample from the distribution of clean images conditioned on both the FBP reconstruction and the incomplete projection data. By explicitly conditioning on the projection data in sampling the clean images, PEDB naturally incorporates data consistency. Extensive experiments demonstrate that PEDB achieves strong performance in CT reconstruction from three types of incomplete data, including sparse-view, limited-angle and truncated projections. For each of these types, PEDB outperforms evaluated state-of-the-art diffusion bridge models across standard, noisy, and domain-shift evaluations. We believe that PEDB has the potential to serve as a general framework for addressing ill-posed CT reconstruction problems and holds promise for future integration into clinical CT systems to facilitate accurate diagnosis.
\appendix

\section{Explanations for Subsection~\ref{sec: image domain diffusion bridge}}
\label{sec:explain preliminary}
In this appendix, we first present the DDBM framework~\citep{zhou2023denoising}, including both the VE and Variance Preserving (VP) schedules, and then explain the simplifications in Subsection~\ref{sec: image domain diffusion bridge}, where only the VE schedule is considered.
\subsection{Denoising Diffusion Bridge Model (DDBM)}
DDBM is designed to sample from $q_{\text{data}}\left(X_0|X_{\text{FBP}}\right)$, the distribution of clean images conditioned on a corrupted image $X_{\text{FBP}}$. Its forward process is defined by applying Doob's $h$-transform to a reference diffusion process. The reference diffusion process is described by the forward SDE:
\begin{equation}
\text{d}X_t=f\left(t\right)X_t\text{d}t+g\left(t\right)\text{d}w,
\label{eq:diffusion_sde}
\end{equation}
where $f\left(t\right)X_t$ is the drift coefficient of $X_t$. Applying Doob's $h$-transform yields the modified forward SDE:
\begin{equation}
\text{d}X_t=\left[f\left(t\right)X_t+g^2\left(t\right)h_t\left(X_t,X_{\text{FBP}} \right)\right]\text{d}t+g\left(t\right)\text{d}w,
\label{eq:full_forward_sde}
\end{equation}
with
\begin{equation}
h_t\left(X_t,X_{\text{FBP}} \right)=\nabla_{X_t}\log p_{T|t}\left(X_{\text{FBP}}|X_t\right),
\label{eq:h_def}
\end{equation}
where $p_{T|t}\left(X_{\text{FBP}}|X_t\right)$ denotes the transition kernel of the reference diffusion process from time $t$ to $T$, evaluated at $X_t$ and $X_{\text{FBP}}$. The function $h_t\left(X_t,X_{\text{FBP}}\right)$ admits the closed-form expression:
\begin{equation}
h_t\left(X_t,X_{\text{FBP}} \right)=\frac{\left(\alpha_t/\alpha_T\right)X_{\text{FBP}}-X_t}{\sigma_t^2\left(\text{SNR}_t/\text{SNR}_T-1\right)},
\label{eq:ptT}
\end{equation}
with parameters
\begin{equation}
\alpha_t=\exp{\left(\int_0^tf\left(\tau \right)\text{d}\tau\right)},\quad\sigma_t^2=\alpha_t^2\int_0^t\frac{g^2\left(\tau\right)}{\alpha_\tau^2}\text{d}\tau,\quad\text{SNR}_t=\frac{\alpha_t^2}{\sigma_t^2}.
\end{equation}
The forward SDE~(\ref{eq:full_forward_sde}) ensures that the forward process almost surely converges to $X_{\text{FBP}}$ at time $t=T$, as given in equation~(\ref{eq:q_T}), and also enables efficient analytical forward sampling:
\begin{equation}
q\left(X_t|X_0,X_{\text{FBP}}\right)=\mathcal{N}\left(X_t;\alpha_t\left(1-\frac{\text{SNR}_T}{\text{SNR}_t}\right)X_0+\frac{\text{SNR}_T}{\text{SNR}_t}\frac{\alpha_t}{\alpha_T}X_{\text{FBP}},\sigma_t^2\left(1-\frac{\text{SNR}_T}{\text{SNR}_t}\right)I\right).
\label{eq:qt|0y}
\end{equation}

In the reverse process, samples from $q_{\text{data}}\left(X_0|X_{\text{FBP}}\right)$ can be generated by initializing with $X_T=X_{\text{FBP}}$ and simulating backward in time according to the reverse SDE:
\begin{equation}
\text{d}X_t=\left[f\left(t\right)X_t+g^2\left(t\right)\left(h_t\left(X_t, X_{\text{FBP}}\right)-\nabla_{X_t}\log q\left(X_t|X_{\text{FBP}}\right)\right)\right]\text{d}t+g\left(t\right)\text{d}\overline{w}.
\label{eq:reverse_sde}
\end{equation}
Under denoising bridge score matching, the score function $\nabla_{X_t}\log q\left(X_t|X_{\text{FBP}}\right)$ has the tractable form:
\begin{equation}
\nabla_{X_t}\log q\left(X_t|X_{\text{FBP}}\right)=\frac{\left(\frac{\text{SNR}_T}{\text{SNR}_t}\frac{\alpha_t}{\alpha_T}X_{\text{FBP}}+\alpha_t\left(1-\frac{\text{SNR}_T}{\text{SNR}_t}\right)D_{\theta^*}\left(X_t,t,X_\text{FBP}\right)\right)-X_t}{\sigma_t^2\left(1-\frac{\text{SNR}_T}{\text{SNR}_t}\right)},
\label{eq: full score}
\end{equation}
where $D_{\theta^*}\left(X_t,t,X_\text{FBP}\right)$ is the image-domain data predictor trained by minimizing the loss in equation~(\ref{eq:loss}).
\subsection{Simplifications in VE Schedule}
The VE schedule corresponds to setting $f(t)=0$, which leads to the following parameter simplifications:
\begin{equation}
\alpha_t=1,\quad\sigma_t^2=\int_0^tg^2\left(\tau\right)\text{d}\tau,\quad\text{SNR}_t=1/\sigma_t^2.
\end{equation}
Under these simplifications, the forward SDE~(\ref{eq:full_forward_sde}) reduces to~(\ref{eq:forward_sde}), the expression for $h_t\left(X_t,X_{\text{FBP}}\right)$ in~(\ref{eq:ptT}) reduces to~(\ref{eq: h}), the expression for the forward sampling distribution~(\ref{eq:qt|0y}) reduces to~(\ref{eq:forward sample}), the reverse SDE~(\ref{eq:reverse_sde}) reduces to~(\ref{eq: image domain reverse sde}), and the expression for the score function~(\ref{eq: full score}) reduces to~(\ref{eq: score matching}).

\section{Proofs}
\subsection{Proofs of Theorem~\ref{theorem1} and Corollary~\ref{corollary1}}
\label{sec: proof_theorem1}
Given both $X_{\text{FBP}}$ and $y$, let $q_t^{\text{forw}}\left(X_t|X_\text{FBP},y\right)$ denote the marginal distribution of the intermediate image $X_t$ obtained by simulating the forward SDE~(\ref{eq:forward_sde}) from $X_0\sim q_{\text{data}}\left(X_0|X_{\text{FBP}},y\right)$ forward to time $t$. Likewise, let $q_t^{\text{back}}\left(X_t|X_\text{FBP},y\right)$ denote the marginal distribution of $X_t$ obtained by simulating the reverse SDE~(\ref{eq: reverse sde}) from $X_T=X_{\text{FBP}}$ backward to time $t$. For Theorem~\ref{theorem1}, our goal is to prove that: 
\begin{equation}
q_t^{\text{forw}}\left(X_t|X_\text{FBP},y\right)=q_t^{\text{back}}\left(X_t|X_\text{FBP},y\right), \quad\forall t\in[0,T].
\label{eq: same_margin}
\end{equation}

We begin by verifying the equality at time $t=T$. Since the forward process almost surely converges to  the corrupted image $X_\text{FBP}$ at time $t=T$, and the reverse process is initialized with $X_T=X_\text{FBP}$, it follows that:
\begin{equation}
q_T^{\text{forw}}\left(X_T|X_\text{FBP},y\right)=q_T^{\text{back}}\left(X_T|X_\text{FBP},y\right)=\delta\left(X_T-X_\text{FBP}\right).
\label{eq: forw_back_T}
\end{equation}

Next, we show that both $q_t^{\text{forw}}\left(X_t|X_\text{FBP},y\right)$ and $q_t^{\text{back}}\left(X_t|X_\text{FBP},y\right)$ satisfy the same time-evolving partial differential equation (PDE). The Fokker-Planck equation~\citep{maoutsa2020interacting} for the forward SDE~(\ref{eq:forward_sde}) describes the time evolution of  $q_t^{\text{forw}}\left(X_t|X_\text{FBP},y\right)$ as:
\begin{multline}
\frac{\partial q_t^{\text{forw}}\left(X_t|X_\text{FBP},y\right)}{\partial t}=-g^2\left(t\right)\nabla_{X_t}\cdot \left(h_t\left(X_t,X_\text{FBP}\right)q_t^{\text{forw}}\left(X_t|X_\text{FBP},y\right)\right)\\+\frac{1}{2}g^2\left(t\right)\Delta_{X_t}q_t^{\text{forw}}\left(X_t|X_\text{FBP},y\right),
\label{eq: forw_fokker}
\end{multline}
where $\Delta_{X_t}=\nabla_{X_t}\cdot\nabla_{X_t}$ is the Laplace operator. Similarly, the Fokker-Planck equation for the reverse SDE~(\ref{eq: reverse sde}) describes the time evolution of  $q_t^{\text{back}}\left(X_t|X_\text{FBP},y\right)$ as:
\begin{multline}
\frac{\partial q_t^{\text{back}}\left(X_t|X_\text{FBP},y\right)}{\partial t}=\\-g^2\left(t\right)\nabla_{X_t}\cdot\left[\left(h_t\left(X_t,X_{\text{FBP}}\right)-\frac{1}{2}\left(1+\gamma^2\right)\nabla_{X_t}\log q_t^{\text{back}}\left(X_t|X_{\text{FBP}},y\right)\right)q_t^{\text{back}}\left(X_t|X_\text{FBP},y\right)\right]\\-\frac{1}{2}\gamma^2g^2\left(t\right)\Delta_{X_t}q_t^{\text{back}}\left(X_t|X_\text{FBP},y\right).
\label{eq: reverse_fokker_raw}
\end{multline}
By substituting the identity
\begin{equation}
\nabla_{X_t}\log q_t^{\text{back}}\left(X_t|X_{\text{FBP}},y\right)=\frac{\nabla_{X_t}q_t^{\text{back}}\left(X_t|X_{\text{FBP}},y\right)}{q_t^{\text{back}}\left(X_t|X_{\text{FBP}},y\right)}
\end{equation}
into equation~(\ref{eq: reverse_fokker_raw}), we obtain:
\begin{multline}
\frac{\partial q_t^{\text{back}}\left(X_t|X_\text{FBP},y\right)}{\partial t}=-g^2\left(t\right)\nabla_{X_t}\cdot \left(h_t\left(X_t,X_\text{FBP}\right)q_t^{\text{back}}\left(X_t|X_\text{FBP},y\right)\right)\\+\frac{1}{2}g^2\left(t\right)\Delta_{X_t}q_t^{\text{back}}\left(X_t|X_\text{FBP},y\right).
\label{eq: back_fokker}
\end{multline}
Comparing equations~(\ref{eq: forw_fokker}) and~(\ref{eq: back_fokker}), we find that both $q_t^{\text{forw}}\left(X_t|X_\text{FBP},y\right)$ and $q_t^{\text{back}}\left(X_t|X_\text{FBP},y\right)$ satisfy the same time-evolving PDE. Together with the shared terminal condition at $t = T$ in equation~(\ref{eq: forw_back_T}), we conclude that the equality in equation~(\ref{eq: same_margin}) holds for all $t \in [0, T] $. This completes the proof of Theorem~\ref{theorem1}.

For Corollary~\ref{corollary1}, we apply the equality in equation~(\ref{eq: same_margin}) at $t=0$:
\begin{equation}
q_0^{\text{back}}\left(X_0|X_\text{FBP},y\right)=q_0^{\text{forw}}\left(X_0|X_\text{FBP},y\right).
\end{equation}
Since the forward process is initialized from $X_0\sim q_{\text{data}}\left(X_0|X_{\text{FBP}},y\right)$, it holds that $q_0^{\text{forw}}\left(X_0|X_\text{FBP},y\right)=q_{\text{data}}\left(X_0|X_{\text{FBP}},y\right)$. Therefore, 
\begin{equation}
q_0^{\text{back}}\left(X_0|X_\text{FBP},y\right)=q_{\text{data}}\left(X_0|X_{\text{FBP}},y\right),
\end{equation}
which completes the proof of Corollary~\ref{corollary1}.

\subsection{Derivation of Equation~(\ref{eq:score})}
\label{sec: direvation_score}
To derive equation~(\ref{eq:score}), we start with the posterior score function $\nabla_{X_t}\log q\left(X_t|X_{\text{FBP}},y\right)$ and proceed as follows:
\begin{equation}
\begin{aligned}
&\nabla_{X_t}\log q\left(X_t|X_{\text{FBP}},y\right)\\&=\frac{1}{q\left(X_t|X_{\text{FBP}},y\right)}\nabla_{X_t}q\left(X_t|X_{\text{FBP}},y\right),\\&=\frac{1}{q\left(X_t|X_{\text{FBP}},y\right)}\nabla_{X_t}\int q\left(X_t|X_0,X_{\text{FBP}},y\right)q_{\text{data}}\left(X_0|X_{\text{FBP}},y\right)\text{d}X_0,\\&=\frac{1}{q\left(X_t|X_{\text{FBP}},y\right)}\nabla_{X_t}\int q\left(X_t|X_0,X_{\text{FBP}}\right)q_{\text{data}}\left(X_0|X_{\text{FBP}},y\right)\text{d}X_0,\\&=\frac{1}{q\left(X_t|X_{\text{FBP}},y\right)}\int \left(\nabla_{X_t}q\left(X_t|X_0,X_{\text{FBP}}\right)\right)q_{\text{data}}\left(X_0|X_{\text{FBP}},y\right)\text{d}X_0
,\\&=\frac{\int \left(\frac{X_{\text{FBP}}}{\overline{\sigma}_t^2}-\frac{\sigma_T^2X_t}{\sigma_t^2\overline{\sigma}_t^2}+\frac{X_{0}}{\sigma_t^2}\right)q\left(X_t|X_0,X_{\text{FBP}}\right)q_{\text{data}}\left(X_0|X_{\text{FBP}},y\right)\text{d}X_0}{q\left(X_t|X_{\text{FBP}},y\right)},\\&=\int \left(\frac{X_{\text{FBP}}}{\overline{\sigma}_t^2}-\frac{\sigma_T^2X_t}{\sigma_t^2\overline{\sigma}_t^2}+\frac{X_0}{\sigma_t^2}\right)\frac{q\left(X_t|X_0,X_{\text{FBP}}\right)q_{\text{data}}\left(X_0|X_{\text{FBP}},y\right)}{q\left(X_t|X_{\text{FBP}},y\right)}\text{d}X_0,\\&=\int \left(\frac{X_{\text{FBP}}}{\overline{\sigma}_t^2}-\frac{\sigma_T^2X_t}{\sigma_t^2\overline{\sigma}_t^2}+\frac{X_0}{\sigma_t^2}\right)\frac{q\left(X_t|X_0,X_{\text{FBP}},y\right)q_{\text{data}}\left(X_0|X_{\text{FBP}},y\right)}{q\left(X_t|X_{\text{FBP}},y\right)}\text{d}X_0,\\&=\int \left(\frac{X_{\text{FBP}}}{\overline{\sigma}_t^2}-\frac{\sigma_T^2X_t}{\sigma_t^2\overline{\sigma}_t^2}+\frac{X_0}{\sigma_t^2}\right)q_{\text{data}}\left(X_0|X_t,X_\text{FBP},y\right)\text{d}X_0,\\&=\frac{1}{\overline{\sigma}_t^2}X_{\text{FBP}}-\frac{\sigma_T^2}{\sigma_t^2\overline{\sigma}_t^2}X_t+\frac{1}{\sigma_t^2}\int X_0q_{\text{data}}\left(X_0|X_t,X_\text{FBP},y\right)\text{d}X_0,\\&=\frac{1}{\overline{\sigma}_t^2}X_{\text{FBP}}-\frac{\sigma_T^2}{\sigma_t^2\overline{\sigma}_t^2}X_t+\frac{1}{\sigma_t^2}\hat{X}_{0,\text{p}}^{\left(t\right)}.
\label{eq: derive_score}
\end{aligned}
\end{equation}
We now explain each equality in the derivation above, with the numbering below matching the corresponding equality in equation~(\ref{eq: derive_score}): (1) apply the chain rule; (2) use the law of total probability to express $q\left(X_t|X_{\text{FBP}},y\right)$ as an integral over $X_0$; (3) use the conditional independence between $y$ and $X_t$ given $X_0$ and $X_{\text{FBP}}$, allowing $q\left(X_t|X_0,X_{\text{FBP}},y\right)=q\left(X_t|X_0,X_{\text{FBP}}\right)$; (4) move the gradient operator inside the integral; (5) compute the gradient of the Gaussian distribution $q\left(X_t|X_0,X_{\text{FBP}}\right)$ with respect to $X_t$;
(6) bring the denominator $q\left(X_t|X_{\text{FBP}},y\right)$ inside the integral; (7) again apply the conditional independence between $y$ and $X_t$ given $X_0$ and $X_{\text{FBP}}$ ; (8) apply Bayes' rule to recognize the distribution $q_{\text{data}}\left(X_0|X_t,X_{\text{FBP}},y\right)$; (9) move terms independent of $X_0$ outside the integral; and (10) use the definition of the projection-embedded expected mean $\hat{X}_{0,\text{p}}^{(t)}$ as given in equation~(\ref{eq: x0c}). This completes the derivation.
\subsection{Proof of Equivalence between Reverse SDEs~(\ref{eq: reverse sde}) and~(\ref{eq: equivalent sde})}
\label{sec: sde_equivalence}
To establish the equivalence between the reverse SDEs~(\ref{eq: reverse sde}) and~(\ref{eq: equivalent sde}), we begin by applying the chain rule and product rule to derive the following differential identities:
\begin{equation}
\text{d}\left(\frac{\sigma_t}{\overline{\sigma}_t}\right)^{1-\gamma^2}=\left(1-\gamma^2\right)\left(\frac{\overline{\sigma}_t}{\sigma_t}\right)^{\gamma^2}\frac{\sigma_T^2}{2\sigma_t\overline{\sigma}_t^3}g^2\left(t\right)\text{d}t,
\label{eq: chain_rule}
\end{equation}
\begin{equation}
\text{d}\left(\frac{\overline{\sigma}_t}{\sigma_t}\right)^{1+\gamma^2}=-\left(1+\gamma^2\right)\left(\frac{\overline{\sigma}_t}{\sigma_t}\right)^{\gamma^2}\frac{\sigma_T^2}{2\sigma_t^3\overline{\sigma}_t}g^2\left(t\right)\text{d}t,
\end{equation}
\begin{equation}
\begin{aligned}
\text{d}\frac{X_t}{\sigma_t^{\left(1+\gamma^2\right)}\overline{\sigma}_t^{\left(1-\gamma^2\right)}}&=\frac{\text{d}X_t}{\sigma_t^{\left(1+\gamma^2\right)}\overline{\sigma}_t^{\left(1-\gamma^2\right)}}+X_t\text{d}\frac{1}{\sigma_t^{\left(1+\gamma^2\right)}\overline{\sigma}_t^{\left(1-\gamma^2\right)}}\\&=\frac{\text{d}X_t}{\sigma_t^{\left(1+\gamma^2\right)}\overline{\sigma}_t^{\left(1-\gamma^2\right)}}+X_t\left(\frac{\overline{\sigma}_t}{\sigma_t}\right)^{\gamma^2}\frac{2\sigma_t^2-\left(1+\gamma^2\right)\sigma_T^2}{2\sigma_t^3\overline{\sigma}_t^3}g^2\left(t\right)\text{d}t.
\label{eq: product_rule}
\end{aligned}
\end{equation}
Substituting equations~(\ref{eq: chain_rule}) through~(\ref{eq: product_rule}) into the reverse SDE~(\ref{eq: equivalent sde}) and multiplying both sides by the factor $\sigma_t^{\left(1+\gamma^2\right)}\overline{\sigma}_t^{\left(1-\gamma^2\right)}$, we obtain the following intermediate form:
\begin{equation}
\text{d}X_t=\left[\frac{\left(1+\gamma^2\right)\sigma_T^2-2\sigma_t^2}{2\sigma_t^2\overline{\sigma}_t^2}X_t+\frac{1-\gamma^2}{2\overline{\sigma}_t^2}X_{\text{FBP}}-\frac{1+\gamma^2}{2\sigma_t^2}\hat{X}_{0,\text{p}}^{\left(t\right)}\right]g^2\left(t\right)\text{d}t+\gamma g\left(t\right)\text{d}\overline{w}.
\label{eq: intermediate_SDE}
\end{equation}

Separately, by substituting the expressions for the linear drift term $h_t\left(X_t,X_{\text{FBP}}\right)$ from equation~(\ref{eq: h}) and the posterior score $\nabla_{X_t}\log\left(X_t|X_{\text{FBP}},y\right)$ from equation~(\ref{eq:score}) into the reverse SDE~(\ref{eq: reverse sde}), we obtain the same intermediate form~(\ref{eq: intermediate_SDE}). 

Since both reverse SDEs~(\ref{eq: reverse sde}) and~(\ref{eq: equivalent sde}) can be analytically transformed into the same intermediate expression~(\ref{eq: intermediate_SDE}), we conclude that they are equivalent. This completes the proof.
\subsection{Derivations of Equations~(\ref{eq: discretize reverse sde}) through~(\ref{eq: eta_t})}
\label{sec: derive_discretize_sde}
To derive equation~(\ref{eq: discretize reverse sde}), we begin by integrating both sides of the reverse SDE~(\ref{eq: equivalent sde}) over the time interval $[t-\Delta t,t]$. This yields:
\begin{multline}
\frac{X_{t-\Delta t}}{\sigma_{t-\Delta t}^{\left(1+\gamma^2\right)}\overline{\sigma}_{t-\Delta t}^{\left(1-\gamma^2\right)}}=\frac{X_{t}}{\sigma_{t}^{\left(1+\gamma^2\right)}\overline{\sigma}_{t}^{\left(1-\gamma^2\right)}}+\frac{X_{\text{FBP}}}{\sigma_T^2}\left(\left(\frac{\sigma_{t-\Delta t}}{\overline{\sigma}_{t-\Delta t}}\right)^{1-\gamma^2}-\left(\frac{\sigma_{t}}{\overline{\sigma}_{t}}\right)^{1-\gamma^2}\right)\\+\frac{\hat{X}_{0,\text{p}}^{\left(t\right)}}{\sigma_T^2}\left(\left(\frac{\overline{\sigma}_{t-\Delta t}}{\sigma_{t-\Delta t}}\right)^{1+\gamma^2}-\left(\frac{\overline{\sigma}_{t}}{\sigma_{t}}\right)^{1+\gamma^2}\right)+\sqrt{\int_{t-\Delta t}^t\frac{\gamma^2g^2\left(\tau\right)}{\sigma_\tau^{\left(2+2\gamma^2\right)}\overline{\sigma}_\tau^{\left(2-2\gamma^2\right)}}\text{d}\tau}\epsilon_t,
\label{eq: integrate_sde}
\end{multline}
where the projection-embedded expected mean $\hat{X}_{0,\text{p}}^{\left(t\right)}$ is assumed to remain approximately constant over the time interval $[t-\Delta t,t]$, and $\epsilon_t \sim \mathcal{N}(0, I)$ is standard Gaussian noise. Next, to evaluate the integral in the last term of equation~(\ref{eq: integrate_sde}), we observe that the antiderivative of the integrand is given by $-\frac{1}{\sigma_T^2}\left(\frac{\overline{\sigma}_\tau}{\sigma_\tau}\right)^{2\gamma^2}$. Thus, we obtain:
\begin{equation}
\int_{t-\Delta t}^t\frac{\gamma^2g^2\left(\tau\right)}{\sigma_{\tau}^{(2+2\gamma^2)}\overline{\sigma}_{\tau}^{(2-2\gamma^2)}}\text{d}\tau=\frac{1}{\sigma_T^2}\left(\left(\frac{\overline{\sigma}_{t-\Delta t}}{\sigma_{t-\Delta t}}\right)^{2\gamma^2}-\left(\frac{\overline{\sigma}_{t}}{\sigma_{t}}\right)^{2\gamma^2}\right).
\end{equation}
Substituting this expression into equation~(\ref{eq: integrate_sde}) and multiplying both sides by the factor $\sigma_{t-\Delta t}^{\left(1+\gamma^2\right)}\overline{\sigma}_{t-\Delta t}^{\left(1-\gamma^2\right)}$, we obtain the discretized update in equation~(\ref{eq: discretize reverse sde}), where the coefficients $a_t$, $b_t$, and $c_t$ are defined in equation~(\ref{eq:ABC}) and the noise scaling factor $\eta_t$ is defined in equation~(\ref{eq: eta_t}). This completes the derivation.

\section{Projection Geometry}
\label{sec: projection geometry}
\begin{table}[t]
\footnotesize
\centering
\setlength{\tabcolsep}{2.2mm}
\begin{tabular}{l|c|c|c|c|c|c}
\toprule
&\multicolumn{3}{c|}{Simulations}&\multicolumn{3}{c}{Real Experiments}\\
\midrule
Source-to-Isocenter &\multicolumn{3}{c|}{595 mm}&\multicolumn{3}{c}{615 mm}\\
Source-to-Detector &\multicolumn{3}{c|}{1086.5 mm}&\multicolumn{3}{c}{1098 mm}\\
Ray Spacing Type &\multicolumn{3}{c|}{Equispaced}&\multicolumn{3}{c}{Equiangular}\\
Detector Pixel Size &\multicolumn{3}{c|}{0.83 mm}&\multicolumn{3}{c}{1.05 mm}\\
Image Size &\multicolumn{3}{c|}{512$\times$512}&\multicolumn{3}{c}{512$\times$512}\\
Image Pixel Size&\multicolumn{3}{c|}{0.5 mm}&\multicolumn{3}{c}{0.625 mm}\\
\midrule
Incomplete Type&\makecell[c]{Sparse\\ View}&\makecell[c]{Limited\\ Angle}&\makecell[c]{Truncated}&\makecell[c]{Sparse\\ View}&\makecell[c]{Limited\\ Angle}&\makecell[c]{Truncated}\\
\midrule
View Count & 120 & 240 & 720 & 120 & 240 & 720 \\
Angular Coverage & 360° & 120° & 360° & 360° & 120° & 360° \\
Detector Pixels& 800 & 800 & 400 & 848 & 848 & 384 \\
\bottomrule
\end{tabular}
\caption{Fan-beam projection geometry settings for simulations and real experiments.}
\label{tab:geometry_settings}
\end{table}
We employed fan-beam projection geometry for both simulations and real experiments. Detailed settings are summarized in Table~\ref{tab:geometry_settings}.
\section{I$^2$SB-Specified and DDBM-Specified Predictors}
\label{sec: specification}

\begin{table}[t]
\footnotesize
\centering
\setlength{\tabcolsep}{0.1mm}
\begin{tabular}{lcc}
\toprule
&I$^2$SB&DDBM (VE)\\
\midrule
\multicolumn{3}{l}{\textbf{Schedule}}\\
$T$&1&$T$\\
$g^2\left(t\right)$&$\left(\frac{\sqrt{\beta_1}+\sqrt{\beta_0}}{2}-\frac{\sqrt{\beta_1}-\sqrt{\beta_0}}{2}\lvert2t-1\rvert\right)^2$&$2t$\\
$\sigma_t^2$&$\int_0^tg^2\left(\tau\right)\text{d}\tau$&$t^2$\\
Parameters&$\beta_0=0.1,\quad\beta_1=0.3$&$T=2.5$\\
\midrule
\multicolumn{3}{l}{\textbf{Training Loss weighting function}}\\
$\lambda\left(t\right)$&$\frac{1}{\sigma_t^2}$&$\frac{w\left(t\right)}{\sigma_t^4}$\\
\midrule
\multicolumn{3}{l}{\textbf{Network Parameterization and Preconditioning}}\\
$D_{\theta}\left(X_t,t,X_\text{FBP}\right)$&$X_t-\sigma_tF_{\theta}\left(X_t,t,X_\text{FBP}\right)$&$c_{\text{skip}}\left(t\right)X_t+c_{\text{out}}\left(t\right)F_{\theta}\left(c_{\text{in}}\left(t\right)X_t,c_{\text{noise}}\left(t\right), X_{\text{FBP}}\right)$\\
\bottomrule
\end{tabular}
\caption{Specifications of the design space within the DDBM framework for the I$^2$SB-specified and DDBM-specified image-domain data predictors. The notation $F_{\theta}$ denotes the neural network to be trained. The definitions of $w(t)$, $c_{\text{skip}}(t)$, $c_{\text{out}}(t)$, $c_{\text{in}}(t)$, and $c_{\text{noise}}(t)$ are consistent with those in the original DDBM ~\citep{zhou2023denoising}.}.
\label{table: specification}
\end{table}
Within the unified DDBM framework presented in Subsection~\ref{sec: image domain diffusion bridge}, both I$^2$SB-specified and DDBM-specified image-domain data predictors are trained by minimizing the loss in equation~(\ref{eq:loss}), but differ in the specifications of the framework’s design space, including the formulations of the diffusion coefficient $g(t)$ and the loss weighting function $\lambda(t)$, along with the network parameterization and preconditioning strategies. We specify these aspects for both predictors in Table~\ref{table: specification}.
\section{Implementation Details for Comparison Methods}
\label{sec: comparison_method}

\begin{table}[t]
\footnotesize
\begin{center}
\setlength{\tabcolsep}{1.72mm}{
\begin{tabular}{lccccccccc}
\toprule
 &&\multicolumn{2}{c}{TV}&\multicolumn{2}{c}{DPS}&DDBM&CDDB&\multicolumn{2}{c}{PEDB} \\
 \cmidrule(r){3-4}\cmidrule(r){5-6}\cmidrule(r){7-7}\cmidrule(r){8-8}\cmidrule(r){9-10}
      &&$\mu$&$\lambda$&NFE& $t_0/T$ &$s$        &$c$ &$k_x^{\dagger}$ &$k_x^{\dagger\dagger}$\\
\midrule
& Standard  & 1000        & 0.1       &50& 0.05        & 0.4        &0.1 & 0         & 0        \\
& Low-Noise  & 1000     & 0.1        & 50    & 0.05     & 0.4       & 0.1  &0.03          & 0      \\
Sparse-View&High-Noise   & 300         &0.3       & 50& 0.05         &0.4       & 0.1&3           & 2                \\
& Domain-Shift& 1000&0.1&20&0.02&0.4&0.1&0&0\\
& Real&300&1&50&0.05&0.6&0.1&0.3&0.3\\
\midrule
& Standard& 3000&0.1&600 & 0.6&0.4&0.03&0&0\\
& Low-Noise&300&0.03&600 &0.6&0.4&0.03&0&0\\
Limited-Angle& High-Noise&100&0.01&600 &0.6&0.33&0.03&0&0\\

&Domain-Shift&3000&0.1&400&0.4&0.4&0.03&0&0\\
&Real&300&1&400&0.4&0.4&0.03&0&0\\
\midrule
&Standard&3000&0.3&300&0.4&0.4&0.01&0&0\\
&Low-Noise&300&0.03&300&0.4&0.6&0.01&0&0\\
Truncated&High-Noise&100&0.03&300&0.4&0.6&0.01&0&0\\
&Domain-Shift&300&0.01&400&0.6&0.33&0.01&0&0\\
&Real&300&0.3&300&0.3&0.4&0.03&0.1&0\\

\bottomrule
\end{tabular}}
\caption{
Inference hyperparameter settings. The hyperparameter notation for TV is adopted from~\citet{goldstein2009split}; for DPS, from~\citet{chung2022diffusion} and~\citet{meng2021sdedit}; for DDBM, from~\citet{zhou2023denoising}; and for CDDB, from~\citet{chung2024direct}. $^{\dagger}$ $k_x$ values used in PEDB with DDBM-specified image-domain data predictors. $^{\dagger\dagger}$ $k_x$ values used in PEDB with I$^2$SB-specified image-domain data predictors.}
\label{tab: parameters}
\end{center}
\end{table}
This appendix specifies the implementation details for all comparison methods, including TV, U-Net, DPS, DDBM, I$^2$SB, and CDDB.

\textbf{TV~\citep{goldstein2009split}.} We used isotropic TV as a regularizer combined with  the data consistency constraint. Optimization was performed using the Split Bregman method~\citep{goldstein2009split} with 1000 iterations. The values of the data consistency weight $\mu$ and the Lagrange multiplier $\lambda$ are listed in Table~\ref{tab: parameters}.

\textbf{U-Net~\citep{ronneberger2015u}.} U-Net models were trained to predict clean images from the corresponding FBP reconstructions using the MSE loss. The network architecture closely followed the 2D residual U-Net used in DDPM~\citep{ho2020denoising}, excluding only the time-conditioning components. All network training configurations were kept the same as those used in PEDB.

\textbf{DPS~\citep{chung2022diffusion}.} An unconditional DDPM model~\citep{ho2020denoising} was trained for 100000 iterations, with all other training configurations the same as those used in PEDB. The network architecture followed the 2D residual U-Net used in DDPM. During inference, the data consistency step size $\zeta_i$ was set to $0.5/\Vert y - A(\hat{x}_0(x_i)) \Vert$. Instead of initializing at time $T$ with pure Gaussian noise, we adopted the jump-start technique from Stochastic Differential Editing (SDEdit)~\citep{meng2021sdedit}, initializing at time $t_0$ using the FBP reconstruction with added Gaussian noise. The values for NFE and $t_0$ are listed in Table~\ref{tab: parameters}.

\textbf{DDBM~\citep{zhou2023denoising}.} We used the same DDBM-specified image-domain data predictors as PEDB. During inference, the NFE was set to 299, and the guidance strength $w$ was set to 1. The values for the step ratio $s$ are listed in Table~\ref{tab: parameters}.

\textbf{I$^2$SB~\citep{liu20232}.} We used the same I$^2$SB-specified image-domain data predictors as PEDB. Inference was conducted with 10 and 50 NFEs.

\textbf{CDDB~\citep{chung2024direct}.} We also used the same I$^2$SB-specified image-domain data predictors as PEDB. Inference was conducted with 10 and 50 NFEs. The values for the data consistency step size $c$ are listed in Table~\ref{tab: parameters}.
\section{Additional Visualization Results}
\label{sec: additional_visual}

\begin{figure*}[!t]
\centering
\includegraphics[width=\textwidth]{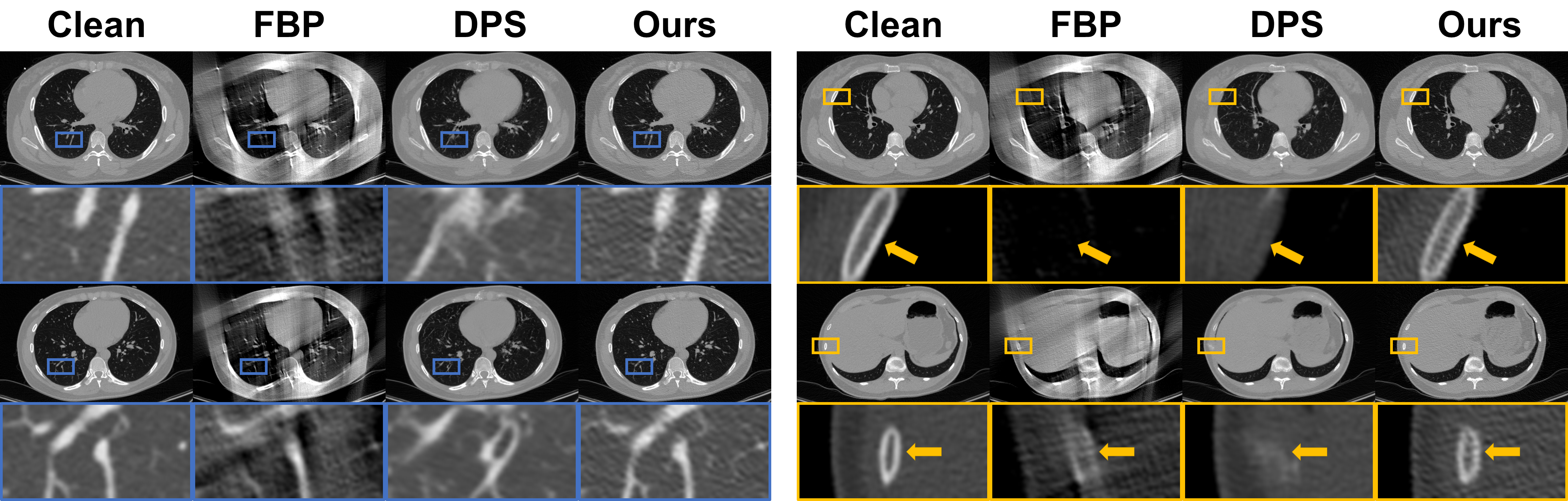}
\caption{Additional visualization results for DPS and our method PEDB in the high-noise simulations for limited-angle CT reconstruction. The regions enclosed by blue and yellow boxes are magnified to highlight structural details. The display window is set to [-1000HU, 800HU] for full chest CT images,  [-1300HU, 200HU] for zoomed-in lung regions, and [-500HU, 1500HU] for zoomed-in bone regions. Our method PEDB uses the I$^2$SB-specified image-domain data predictor with 50 NFEs.}
\label{fig: limit_dps}
\end{figure*}
In Figure~\ref{fig: limit_dps}, we present additional visualization results comparing DPS and our method PEDB in the high-noise simulations for limited-angle CT reconstruction. Compared to DPS, PEDB reconstructs images with higher noise levels but introduces fewer hallucinated details and more accurately restores anatomical structures. Specifically, in the zoomed-in regions enclosed by the blue boxes, DPS generated pulmonary vessels that are inconsistent with those in the clean images, whereas PEDB correctly recovered their anatomical structures. In the regions enclosed by the yellow boxes, DPS completely failed to reconstruct the ribs, while PEDB successfully recovered them.
\section{Inference Time}

\label{sec: time}
\begin{table}[t]
\footnotesize
\begin{center}
\setlength{\tabcolsep}{1.45mm}{
\begin{tabular}{lcccccccc}
\toprule
 &&TV&U-Net&DPS&DDBM&I$^2$SB$^{\dagger}$&CDDB$^{\dagger}$&PEDB$^{\dagger}$ \\

\midrule
& Standard  & 176.4        & 0.2       &23.9&60.1        &2.0         &2.9 &19.2                 \\
& Low-Noise  & 176.4        & 0.2      &23.9&60.1        & 2.0        &2.9 &7.0        \\
Sparse-View&High-Noise   &  176.4       & 0.2     &23.9& 60.1       &2.0         &2.9 & 5.3         \\
& Domain-Shift& 176.4        & 0.2    &9.4& 60.1        &2.0         &2.9 &19.2\\
& Real& 182.6        & 0.2     &23.9& 60.1      &2.0        &2.9 &17.5\\
\midrule
& Standard& 333.8        &  0.2    &340.9& 60.1      &2.0         &3.7 &36.7\\
& Low-Noise& 333.8        &  0.2     &340.9& 60.1        & 2.0        &3.7 &12.0\\
Limited-Angle& High-Noise& 333.8        & 0.2      &340.9& 60.1      &2.0         &3.7 &8.7\\

&Domain-Shift& 333.8        & 0.2    &227.2&  60.1      & 2.0        &3.7 &36.7\\
&Real&   329.7      & 0.2    &226.4&  60.1      & 2.0        &3.7 &32.8\\
\midrule
&Standard& 784.0        & 0.2      &239.9& 60.1       & 2.0        &6.1 &84.3\\
&Low-Noise& 784.0        & 0.2      &239.9& 60.1    & 2.0        &6.1 &25.7\\
Truncated&High-Noise& 784.0        & 0.2   &239.9&60.1       &2.0         &6.1 &17.8\\
&Domain-Shift& 784.0        & 0.2    &320.2& 60.1        & 2.0        &6.1 &84.3\\
&Real& 744.9       &0.2     &233.7& 60.1     &2.0         &5.9 &72.2\\

\bottomrule
\end{tabular}}
\caption{
Inference time per image for all evaluated methods, reported in seconds. $^\dagger$ For I$^2$SB, CDDB, and PEDB, the reported times correspond to NFE = 10; setting NFE to 50 results in approximately five times longer inference times.}
\label{tab: time}
\end{center}
\end{table}
We report the inference times for all evaluated methods in Table~\ref{tab: time}. All experiments were conducted on a single NVIDIA A100-SXM4-40GB GPU. Across different types of incomplete data, variations in inference time for TV, DPS, CDDB, and PEDB arise from differences in the scale of the system matrix. For the same type of incomplete data under different evaluation scenarios, variations in inference time for DPS are caused by different NFE settings, and for PEDB by different numbers of CG iterations.

\section*{Declaration of generative AI and AI-assisted technologies in the manuscript preparation process}
During the preparation of this work the authors used ChatGPT in order to improve language and readability. After using this tool, the authors reviewed and edited the content as needed and take full responsibility for the content of the published article.
\section*{Acknowledgements}
This work was support by National Natural Science Foundation of China (No. 12327809) and Samsung Electronics Co., Ltd.

\end{document}